\documentclass[runningheads]{llncs}

\usepackage[T1]{fontenc}

\usepackage{graphicx}
\usepackage{comment}
\usepackage{amsmath,amssymb}
\usepackage{color}
\usepackage{url}
\usepackage{hyperref}

\usepackage[numbers]{natbib}
\usepackage{booktabs}
\usepackage{xspace}
\usepackage{xcolor, colortbl}
\usepackage{subcaption}
\usepackage{amsmath}
\usepackage{cleveref}
\crefname{figure}{Fig.}{Figs.}
\Crefname{figure}{Fig.}{Figs.}
\crefname{table}{Tab.}{Tabs.}
\Crefname{table}{Tab.}{Tabs}
\crefname{section}{Section}{Sections}
\Crefname{section}{Section}{Sections}
\usepackage{multirow}

\newcommand{\multiObjRatioImg}{\textit{Two-object-in-image-probability}\xspace}
\newcommand{\multiObjRatioCap}{\textit{Two-object-in-caption-probability}\xspace}
\newcommand{\attrsInCap}{\textit{Attributes-per-object-in-caption}\xspace}
\newcommand{\saliencyBias}{\textit{Saliency bias}\xspace}

\makeatletter
\def\thanks#1{\protected@xdef\@thanks{\@thanks
        \protect\footnotetext{#1}}}
\makeatother

\begin{document}

\title{Common Data Properties Limit Object-Attribute Binding in CLIP}

\author{Bijay Gurung$^{1,2}$   \hspace{0.5em}
David T.~Hoffmann$^{\star,1}$ 
\hspace{0.5em}
Thomas Brox$^{\star,1}$
}
\thanks{$^\star$ Shared last authorship. Correspondence to: gurungb@cs.uni-freiburg.de. \\Code available at: https://github.com/bglearning/data-properties-clip-binding}
\authorrunning{B.~Gurung et al.}
\titlerunning{Common Data Properties Limit Object-Attribute Binding in CLIP}

\institute{
$^{1}$University of Freiburg \quad $^{2}$deepset
}

\maketitle

\begin{abstract}
Contrastive vision-language models like CLIP are used for a large variety of applications, such as zero-shot classification or as vision encoder for multi-modal models.
Despite their popularity, their representations show major limitations.
For instance, CLIP models learn bag-of-words representations and, as a consequence, fail to distinguish whether an image is of ``a yellow submarine and a blue bus'' or ``a blue submarine and a yellow bus''.
Previous attempts to fix this issue added hard negatives during training or modified the architecture, but failed to resolve the problem in its entirety.
We suspect that the missing insights to solve the binding problem for CLIP are hidden in arguably the most important part of learning algorithms: the data.
In this work, we fill this gap by rigorously identifying the influence of data properties on CLIP's ability to learn binding using a synthetic dataset. 
We find that common properties of natural data such as low attribute density, incomplete captions, and the saliency bias, a tendency of human captioners to describe the object that is ``most salient'' to them, have a detrimental effect on binding performance.
In contrast to common belief, we find that neither scaling the batch size, i.e., implicitly adding more hard negatives, nor explicitly creating hard negatives enables CLIP to learn reliable binding.
Only when the data expresses our identified data properties does CLIP learn almost perfect binding.

\end{abstract}

\section{Introduction}

Contrastive Vision-Language models 
like CLIP~\cite{clip} are a cornerstone of computer vision.
CLIP-like models are used for various applications, ranging from zero-shot classification to text-to-image and image-to-text retrieval. 
The image encoders of CLIP-like models are a standard choice for the vision encoder in multi-modal language models~\cite{beyer2024paligemma,chen2023pali} and the text encoders have been used to guide image generation~\cite{ nichol2021glide, ramesh2022hierarchical,rombach2022high}.
To mitigate deficiencies of CLIP, a whole family of variants with minor architectural differences or modifications of the loss function have been proposed~\cite{oc-clip,openclip,declip,MetaCLIP,siglip,zhong2022regionclip}, which we will refer to as CLIP-like models.

\begin{figure}[t!]
    \centering
    \begin{subfigure}{0.28\textwidth}
\includegraphics[trim={0, 3.75cm, 0, -60pt}, clip, width=1\linewidth]{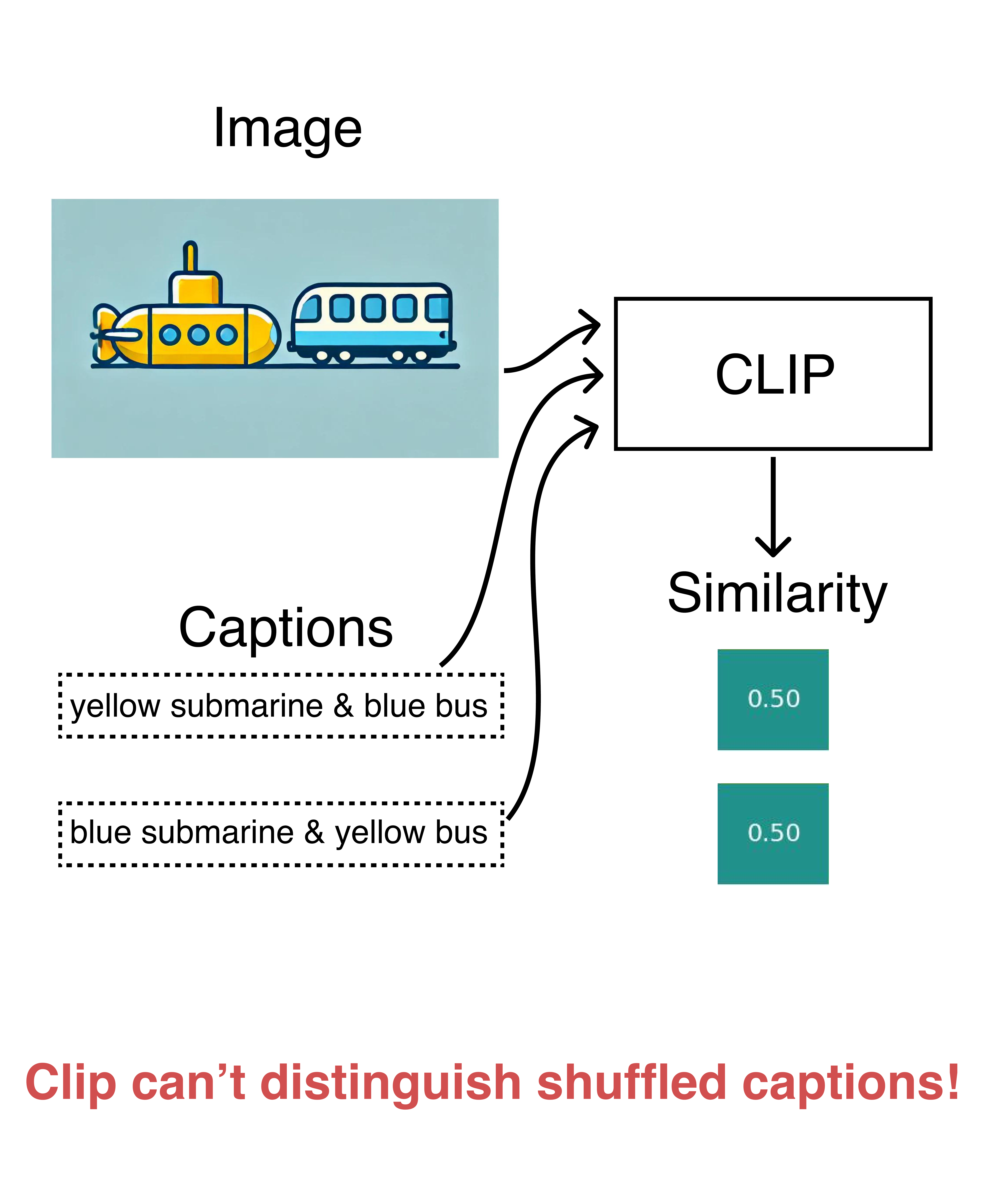}
        \vspace{-2\baselineskip}
        \caption{}
        \label{fig:clip-binding-example}
    \end{subfigure}    
    \begin{subfigure}{0.38\textwidth}    
        \includegraphics[width=1\linewidth]{figures/teaser-fig-b.pdf}
        \vspace{-2\baselineskip}
        \caption{}
        \label{fig:data-setup-poor}
    \end{subfigure}    
    \begin{subfigure}{0.30\textwidth}    
        \includegraphics[width=1\linewidth]{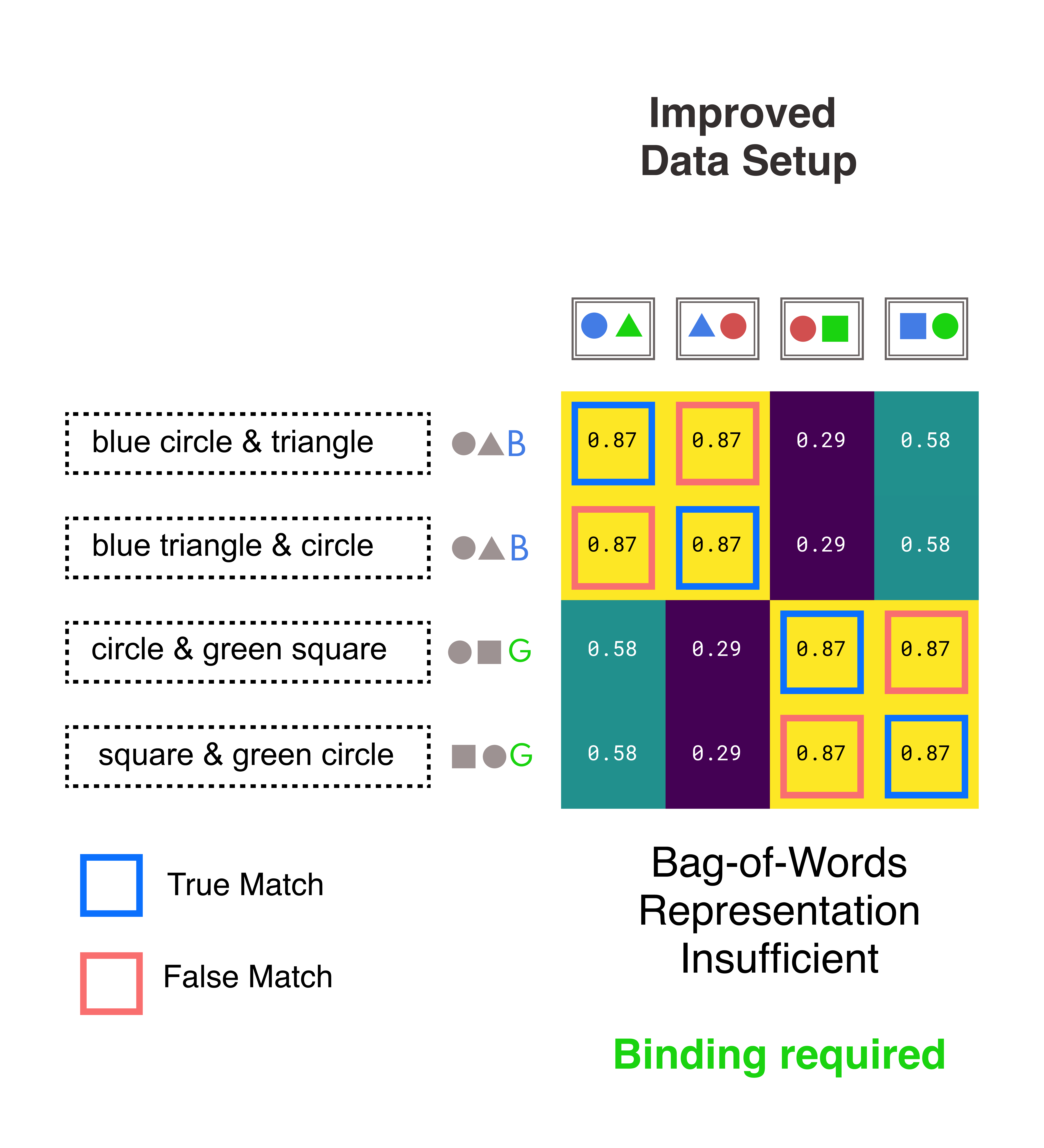}
        \vspace{-2\baselineskip}
        \caption{}
        \label{fig:data-setup-improved}
    \end{subfigure}    
   \vspace{-2mm}
    \caption{\textbf{(a)} The object-attribute binding problem: CLIP can't distinguish between two captions with swapped attributes. \textbf{(b)} We identify data properties that contribute to poor object-attribute binding. If 1) too few or too many attributes per object are in the caption, or 2) too few images with multiple objects or 3) caption mentions too few objects, 4) captioners focus more on salient objects, a bag-of-words (BOW) representation is sufficient to find the correct sample within a batch.
Here we study how these data properties influence object-attribute binding of CLIP models and find \textbf{(c)} setups that lead to robust binding accuracy of CLIP models, suggesting the binding problem of CLIP is a data problem.}
  \vspace{-3mm}
    \label{fig:overall-summary}
\end{figure}

Despite the great success and broad applicability of CLIP-like models, they unfortunately fail at the fundamental task of object-attribute binding~\cite{clip_bind_concept,lemonsPurple,clip_bow}. 
For instance, as shown in~\cref{fig:clip-binding-example} CLIP can't distinguish the ``yellow submarine and blue bus" and the incorrect caption in which the color is swapped.
This is linked to the finding by Yuksekgonul et al.~\citep{clip_bow} that CLIP models represent images and text in a bag-of-words (BOW) representation, i.e.,~CLIP appears to extract a set of concepts from an image and represents an image by an unordered set of these concepts.
Naturally, a BOW representation is, for many tasks, insufficient, can result in unexpected results, and can even cause security risks. 
For example, a self-driving car using BOW representation might fail to bind \textit{red} to the traffic light and confuse a \textit{green}-shirted pedestrian with a green traffic light.

Multiple previous works tried to resolve CLIP's inability to learn reliable object-attribute binding. 
For instance, Yuksekgonul et al.~\citep{clip_bow} claim that CLIP fails to learn binding because a BOW representation is sufficient to associate each caption to the correct image.
As a remedy, they propose to mine hard negatives.
However, as revealed later in Hsieh et al.~\citep{hsieh2023sugarcrepe}, CLIP's binding ability increases only marginally on a more general test set.
Others assume that CLIP's architecture limits its ability to learn object-attribute binding and try to fix the problem by adding an object-centric learning inspired cross-modal interaction layer~\cite{oc-clip}.
But these changes are also insufficient to solve the problem entirely.

This leads to the central question: What prevents CLIP from learning binding? Is it the poor scaling properties of contrastive losses in a large combinatorial space, as indicated by Yuksekgonul et al.~\citep{clip_bow}, or architectural limitations as proposed by Assouel et al.~\citep{oc-clip}? 
We suspect that neither fully explains it. 
Previous works focused on the loss and architecture, but missed arguably the most important component of each learning algorithm: the data. 
Indeed, as shown in \cref{fig:data-setup-poor}, we find various data properties common in image-caption datasets to promote BOW representations, as they make the learning of binding obsolete.
Could properties of the data itself limit the binding of CLIP models?
Could better filtering or re-captioning help? And what would ideal data look like?

Unfortunately, studying the influence of data properties is nontrivial for web-scraped data: the properties are unknown and annotating them is too expensive. 
Without this annotation it is impossible to run clean data interventions.
Even with these annotations, it is necessary to train many CLIP models on different settings, which quickly becomes too expensive.
As a remedy, we define a synthetic data generating process that gives us full control over the data properties.
To ensure the data properties of our synthetic dataset are representative of real data we annotate the data properties for 100 images from CC12M~\cite{changpinyo2021cc12m} and pick our default setup close to those observed on real data. 
Diverging from this ``realistic'' setup, we can study the influence of individual properties.

But can we be sure that relationships found on synthetic data transfer to real data?
Note that studying learning algorithms in a controlled setup reveals basic properties and relations of this learning algorithm.
We do not expect that the learning algorithm changes its behavior fundamentally in larger scale training on more complex data. 
A remaining caveat is that the relations found in our setup could be overshadowed by other factors on real data. Even then this work still contributes to a better understanding of CLIP models.

In summary, our contributions are the following: 
1) We design a clean synthetic dataset which allows us to study object-attribute binding of CLIP models in a controllable full-information setup.
In particular, we analyze the influence of the \textbf{ratio of multi-object to single-object images}, the influence of \textbf{number of objects described in the caption}, the \textbf{number of attributes per object in the caption}, and the influence of \textbf{saliency bias} of annotators on object-attribute binding of CLIP models.
2) We find that the number of objects in the image and in the caption have a significant influence on binding. We find an inverse u-shaped relation between attributes in the caption and binding accuracy, i.e., too few and too many attributes in the caption both hurt binding. We further find that the saliency bias of annotators might be a key factor limiting object-attribute binding. 
3) We confirm that our findings are not due to flaws in the experimental design by scaling up the batch size and model complexity. 
Furthermore, we observe that the trends persist out-of-distribution, ruling out that our models learned a shortcut.
4) Finally, we show that hard negatives alone are insufficient to learn strong binding, and that the data affects the binding accuracy more.

\section{Related Works}
Yuksekgonul et al.~\cite{clip_bow} show that CLIP learns a bag-of-words (BOW) representation and hypothesize this is due to batches lacking hard negatives. Without them, the contrastive loss can be minimized with a BOW representation which CLIP learns (following a simplicity bias).
Similarly, Tang et al.~\cite{lemonsPurple} identify Concept Association Bias (CAB), where CLIP learns strong associations between highly correlated factors (e.g.~``purple'' and ``eggplant'') also leading to binding failures.
 
Several approaches aimed to improve CLIP's binding performance~\cite{oc-clip, doveh2023dense, clip_bow}.
Yuksekgonul et al.~\cite{clip_bow} propose NegCLIP, which is fine-tuned on hard negatives constructed by swapping attributes, object, and relations in the caption.
However, later works revealed that the success was overestimated, due to overfitting to specific types of binding used for testing~\citep{hsieh2023sugarcrepe}.
Assouel et al.~\citep{oc-clip} propose cross-modal interactions, inspired by object-centric learning, and LLM-based caption decomposition. Despite outsourcing most of the binding to an LLM their approach only partially solves the binding problem, as can be seen by the reported binding-accuracy on Sugarcrepe~\cite{hsieh2023sugarcrepe}.
Enriching the captions with LLMs also leads to some improvements~\citep{doveh2023dense,doveh2023teaching}, as does generating hard negatives via in-context learning~\cite{patel2024tripletclip}, 
however, these methods also do not solve the binding problem entirely.
Koishigarina et al.~\cite{clip_unimodal_bow} argue that the binding information is present unimodally and is lost during the cosine similarity computation and propose a contrastively-trained transformation matrix for alignment, showing partial improvements. However, their study centers on overly simplistic datasets.

The datasets used to study object-attribute binding in a controlled setting are based on CLEVR~\cite{clevr_dataset2017} in \cite{clip_bind_concept} and PUG:SPAR~\cite{pugspar} in \cite{clip_unimodal_bow}.
However, they are limited by their low complexity.
Note, that both datasets are rendered images, however, complexity for the object-attribute binding does not arise from image quality and realism. Instead, the number of attributes and their combinations with objects is the key driver of the binding task complexity.
For instance, PUG:SPAR only has one attribute type (color/material). CLEVR~\cite{clevr_dataset2017} only has 3 object types with 3 attributes, 2 of which have only 2 possible values.
As such, these datasets are not sufficiently complex and do not allow for enough control over their data properties to systematically study the influence of training data on CLIP's binding performance.
For evaluation of models trained on large scale image-caption datasets benchmarks from the Crepe family \cite{sugarcrepe++,hsieh2023sugarcrepe,crepe2023} are commonly used. 
Unfortunately, they do not provide the required control for a systematic analysis of the influence of data properties on binding performance.

In contrast to prior work, we take a data-centric approach to study CLIP's binding problem in detail and develop a synthetic dataset that enables systematic investigation of how common properties of web-scraped image-caption data influence CLIP binding performance. We demonstrate that these data properties of the training data significantly impact binding, and that CLIP trained under favorable data conditions achieves high binding performance.
\section{Experimental Setup}

To systematically study the influence of training data properties on object-attribute binding we require precise control over data characteristics, which natural datasets cannot provide. As a remedy, we create a synthetic dataset that enables explicit manipulation of these data properties. 

\subsection{MADMAN: Fully Controllable Synthetic Dataset}

\begin{figure}[t]
    \centering
    \includegraphics[width=1.0\linewidth]{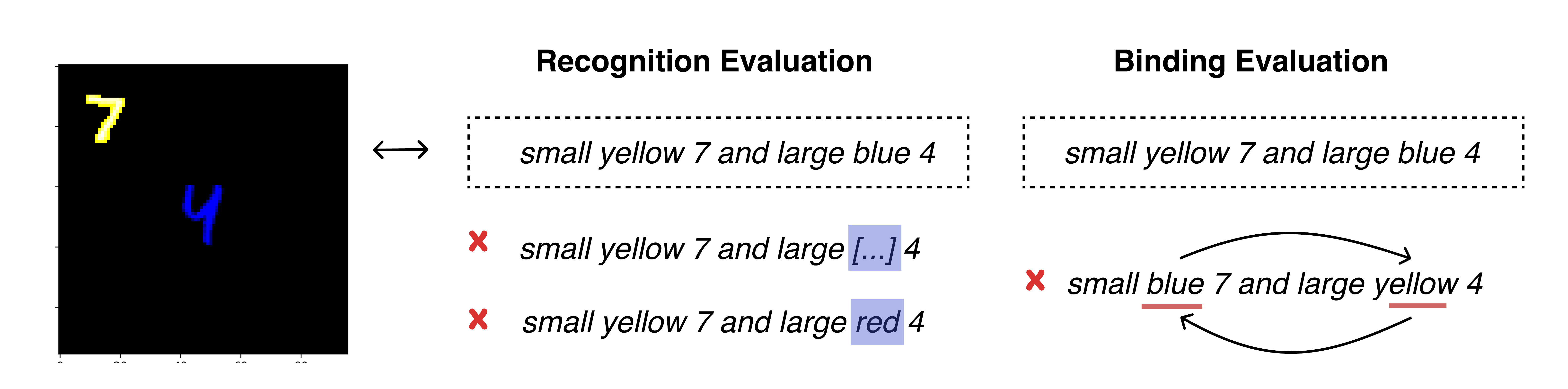}
    \caption{\textbf{MADMAN example and evaluation protocols.} To evaluate \textbf{recognition}, we create zero-shot classification prompts by changing the target attribute. To evaluate \textbf{binding}, we swap the target attribute to create a negative and perform zero-shot classification with the true and negative caption. In practice, we evaluate on captions with 3 and 4 attributes per object.}
    \label{fig:example_recognition_binding}
\end{figure}

We show an example of our dataset Multi-Attribute and Digit for Multi-Attribute biNding (MADMAN) in \cref{fig:example_recognition_binding} and in the Appendix, \cref{creating_madman}. It is designed to study object-attribute binding in a controlled way.
A dataset for this purpose needs to entail images with multiple objects, each with a set of attributes, all of which are described in a caption.
We use the MAD \citep{twoEffects} dataset as the starting point, add support for multiple objects per image, and extend it with \textit{rotation}.
For multi-object images, we create a 3x3 black grid, and fill 2 random cells with images. 
\textbf{Captions} are created by chaining object classes with their attributes, using ``and'' to separate the objects
(see Appendix, \cref{creating_madman} for a complete list of augmentations and examples from our dataset).

\subsubsection{Data properties studied with MADMAN.}
\label{sec:exp_set_dataproperties}
To study the influence of data properties on binding, we create different versions of MADMAN controlling: \textbf{1) \multiObjRatioImg} $p(\text{two-obj-img})$, which controls the probability of a sample to contain 2 objects, \textbf{2) \multiObjRatioCap} 
$p(\text{two-obj-cap} |  \text{two-obj-img}=\text{True})$,
which controls the probability that both objects are described in the caption. 
Note that this is a conditional probability (given the image is a two-object image) and not the prior probability. However, for most experiments it is identical, as we use $p(\text{multi-obj-img})=1$ in our base setup.
We also study \textbf{3)} the influence of the number of \textbf{\attrsInCap}.
We observe that humans mention only a few attributes per object in a caption (see Appendix, \cref{realistic-value-procedure}). 
To model this, we change the probability of an attribute being included in the caption. 
To avoid overfitting to a specific caption length we define probability distributions over the number of attributes per object, and use the expected value to distinguish them (see Appendix, \cref{fig:attr-dists} to see the underlying distributions).
Finally, we are interested in the effect of \textbf{4) \saliencyBias}.
Saliency describes the properties of ``parts popping out'' to human attention.
Images commonly contain a number of salient objects, which are more likely to be described in the caption. 
We model it via a center bias, i.e., we define an object to be salient if it is in the center of the image. Furthermore, salient objects in our setup are always mentioned in the caption and are mentioned first.
We control $p(\text{saliency})$ which refers to the probability of an image to contain a salient object.

\noindent\textbf{Determining realistic parameters for the data properties.}
To draw conclusions about natural data from synthetic data, we need a MADMAN version that has similar data properties as natural data.
To this end,
we annotate the four data properties defined above for 100 images from CC12M \cite{changpinyo2021cc12m}.
For details on the annotation, Appendix~\ref{realistic-value-procedure}.
We refer to this MADMAN variant as ``Realistic''.

\noindent\textbf{Defining Out-of-Distribution.}
To evaluate whether our model learns a compositional representation we need an out-of-distribution (OOD) test set.
We define the following combinations as OOD:
\begin{itemize}
\item[] color: $\{green, red\} \times \{0, 3\}$, $\{blue, magenta\} \times \{4, 5\}$
\item[] scaling: $\{large\} \times \{3, 7\}$, $\{small\} \times \{4, 9\}$
\end{itemize}
Here, $\times$ denotes the Cartesian product. These combinations are included \textbf{only} in our OOD testset (only \cref{ood-results}). The remaining combinations are used for training and our standard test-set.

\subsection{Evaluation and Training Setup}\label{eval-and-training-setup}

The evaluation images always show two objects and both are described in the caption. 
The evaluation captions have 3 attributes per object in 50\% of samples and 4 for the remaining samples.
This ensures a fair comparison across different \attrsInCap settings, as all of them have the same probability mass for 3 and 4 attributes (see \cref{fig:attr-dists}).
We provide an overview of our recognition and binding evaluation in \cref{fig:example_recognition_binding}.

\noindent\textbf{Recognition.}
Our final goal is to evaluate object-attribute binding, however, a prerequisite for binding is that the attribute in question can be recognized. 
We evaluate recognition accuracy per object and attribute.
For each object $o$ and attribute $a$, we use the evaluation captions and vary only $a$ (or $o$) over all  possible values to create false captions. Recognition is evaluated as zero-shot accuracy over the set of these false captions and the ground-truth caption.

\noindent\textbf{Object-Attribute Binding Accuracy.}
To measure object-attribute binding accuracy for an attribute $a$ of the set of attributes $A$ we:
\begin{enumerate}
    \item Filter out \textbf{attributes} $a$ for which recognition is close to or below $p_{\mathrm{chance}}$
    \item Filter out all \textbf{samples} for which attribute $a$ is not recognized for both objects
    \item Evaluate \textit{binding-accuracy} for the remaining samples
\end{enumerate}
Criterion 1) is necessary, as the attribute binding metric has no meaning if the attribute is not even recognized. 
To filter out cases where performance is slightly above chance, we use a  slightly stricter threshold: $1.1 \times p_{\mathrm{chance}}$.
Criterion 2) factors out failures of binding that are caused by a failed recognition.
Finally, for the remaining samples we compute \textit{binding-accuracy}~\cite{hsieh2023sugarcrepe,clip_bow}:
\begin{equation}
    \text{binding-accuracy}(a) = \frac{1}{N} \sum_{n=1}^N  p(c_n^{\text{gt}})>p(c_n^{\text{swap}_i}),
\end{equation}
where $c_n^{\text{gt}}$ is the ground truth caption $c$ of sample $n$, $c_n^{\text{swap}}$ is derived from $c_n^{\text{gt}}$ by swapping attribute $a$ between the objects and $p(c)$ indicates the probability the model assigns to caption $c$ belonging to the image. As this is a two-alternative forced choice task, the chance level is 0.5.

\subsubsection{Model training setup.}
For all results reported in \cref{sec:results}, we train small CLIP models from scratch.
We train each model with three random seeds and report/plot the mean and the 95\% confidence interval.
We vary the data properties of MADMAN as detailed for each experiment. 

Compared to natural datasets, MADMAN exhibits only a few factors of variation. Both the number of object classes and attributes are limited. 
Accordingly, we adjust our model and batch size to match this smaller scale.
Text and image encoders are small transformers with 6 layers and 4 attention heads, each.
We use a batch size of 16 and an embedding size of 32. 
In particular the choice of the batch size plays a critical role in contrastive learning models, as it directly affects the likelihood of sampling hard negatives within a batch. This, in turn, influences the number of distinct features (like binding) the model must learn to minimize the loss. 
We validate our choices in \cref{validating_the_setup}, demonstrating that the observed trends are robust to variations in the specific configuration.

\section{Results}
\label{sec:results}

\begin{table}[t]
    \centering
    \caption{\textbf{Object-Attribute binding in realistic and ideal setup.} Here, Realistic Data refers to a setup with properties determined from real data. \colorbox{blue!18}{Ideal} Data refers to the best performing dataset properties as found in \cref{sec:what-limits}. We ablate the Realistic and Ideal data setting with small and \colorbox{blue!18}{Ideal} (large) batch size and model size. It can be seen that ideal data is more important for \colorbox{green!18}{high binding}. Thus, data properties limit binding, not model size or batch size.}
    \label{tab:teaser_tab}
    \begin{tabular}{lcc|cccccc}
        \toprule
        \multicolumn{3}{c|}{\textbf{Setup}} & \multicolumn{6}{c}{\textbf{Binding Accuracy}} \\
        \cmidrule(lr){1-3} \cmidrule(l){4-9}
        \textbf{Data} & \textbf{Batch} & \textbf{Embed} & \textbf{Color} & \textbf{Scaling} & \textbf{Fracture} & \textbf{Rotation} & \textbf{Swelling} & \textbf{Thickness} \\ 
        \midrule
        Realistic & 16 & 32 & 50.47 & 54.47 & 52.34 & 50.86 & 53.26 & 50.05 \\
        Realistic & 16 & \cellcolor{blue!18}256 & 50.64 & 59.72 & 50.81 & 51.71 & 52.23 & 53.50 \\
        Realistic & \cellcolor{blue!18}256 & 32 & 51.06 & 50.47 & 50.91 & 53.55 & 54.69 & 52.44 \\
        Realistic & \cellcolor{blue!18}256 & \cellcolor{blue!18}256 & 54.74 & 65.47 & 50.13 & 73.96 & 54.61 & 55.36 \\
        \midrule
        \cellcolor{blue!18}Ideal & 16 & 32 & \cellcolor{green!18}94.66 & \cellcolor{green!18}91.28 & 61.93 & \cellcolor{green!18}92.44 & 66.43 & \cellcolor{green!18}90.02 \\
       \cellcolor{blue!18}Ideal & 16 & \cellcolor{blue!18}256 & 88.03 & \cellcolor{green!18}96.99 & 60.17 & \cellcolor{green!18}98.42 & 69.36 & \cellcolor{green!18}95.72 \\
        \cellcolor{blue!18}Ideal & \cellcolor{blue!18}256 & \cellcolor{blue!18}256 & \cellcolor{green!18}100.0 & \cellcolor{green!18}98.48 & \cellcolor{green!18}98.18 & \cellcolor{green!18}99.93 & \cellcolor{green!18}98.64 & \cellcolor{green!18}99.43 \\
        \bottomrule
    \end{tabular}
\end{table}

\subsection{CLIP Only Learns Object-Attribute Binding with Ideal Data}

Can CLIP even learn object-attribute binding? Does the architecture and loss prevent CLIP from learning a non Bag-of-words (BOW) representation? Is maybe a large enough batch size sufficient to encourage CLIP to learn binding via hard negatives?
We can explore these questions using MADMAN with relatively small batch sizes since the number of factors of variation is low.

As can be seen in the top part of \cref{tab:teaser_tab}, neither training with very large batch size, nor training a large model, nor their combination enables binding.
We conclude that CLIP models are unlikely to learn binding when training on data that exhibits realistic values for the four data properties (defined in \cref{sec:exp_set_dataproperties}).
So either CLIP models can not learn binding, or properties of the data inhibit it. 
To find out which of these two explanations is true,
we optimize the training data setup by changing the four data properties. 
As detailed in \cref{sec:what-limits}, we find that changing the data setup has a significant influence on binding accuracy, leading to almost perfect binding (see \cref{tab:teaser_tab}).
We conclude that CLIP models are able to learn object-attribute binding but properties of the data inhibit this. 

\subsection{What Limits the Object-Attribute Binding in CLIP Models?} \label{sec:what-limits}
The previous section shows that CLIP models can learn object-attribute binding under optimal data conditions.
But which properties of the data are important? How do they individually influence the binding? Can we find factors that we could change on real data?
To answer these questions, we individually investigate the influence of the four data properties defined in \cref{sec:exp_set_dataproperties} on the binding performance.
Unless otherwise mentioned, we set \multiObjRatioImg to $1.0$,
\multiObjRatioCap to $1.0$, the expected number of \attrsInCap to $1.8$ and \saliencyBias to 0.
For all experiments below, we vary one data property and keep the others constant. By varying one property we create different variants of MADMAN. For each of these versions we train 3 CLIP models from scratch and report the average binding accuracy.

\subsubsection{What's the influence of the ratio of multi-object images on binding?}
\begin{figure}[t]
    \centering
    \begin{subfigure}[b]{0.49\textwidth}
    
        \includegraphics[width=1\linewidth]{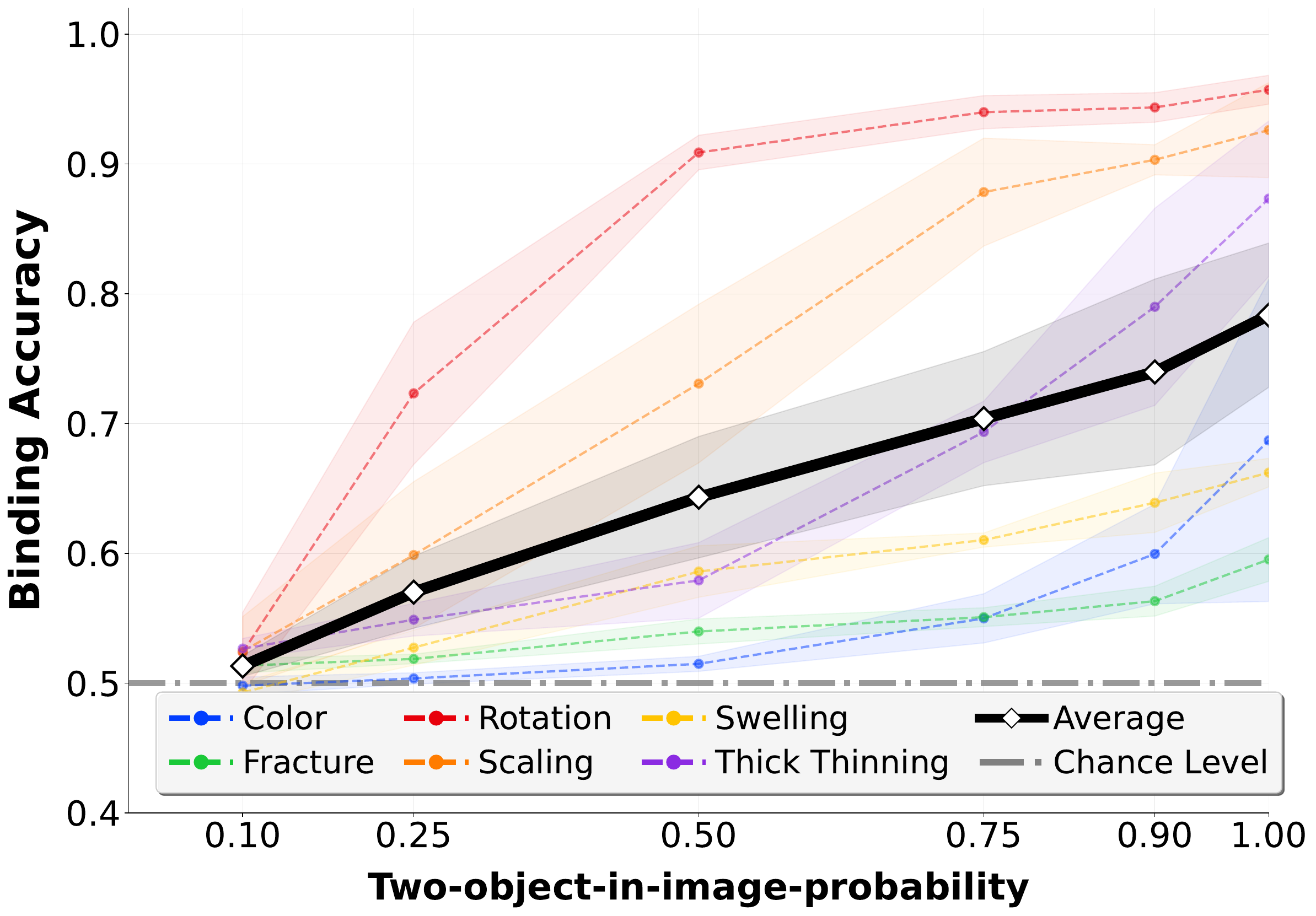}
        \caption{\textbf{\multiObjRatioImg}}
        \label{fig:multi-object_image_ratio}
    \end{subfigure}    
    \begin{subfigure}[b]{0.49\textwidth}    
        \includegraphics[width=1\linewidth]{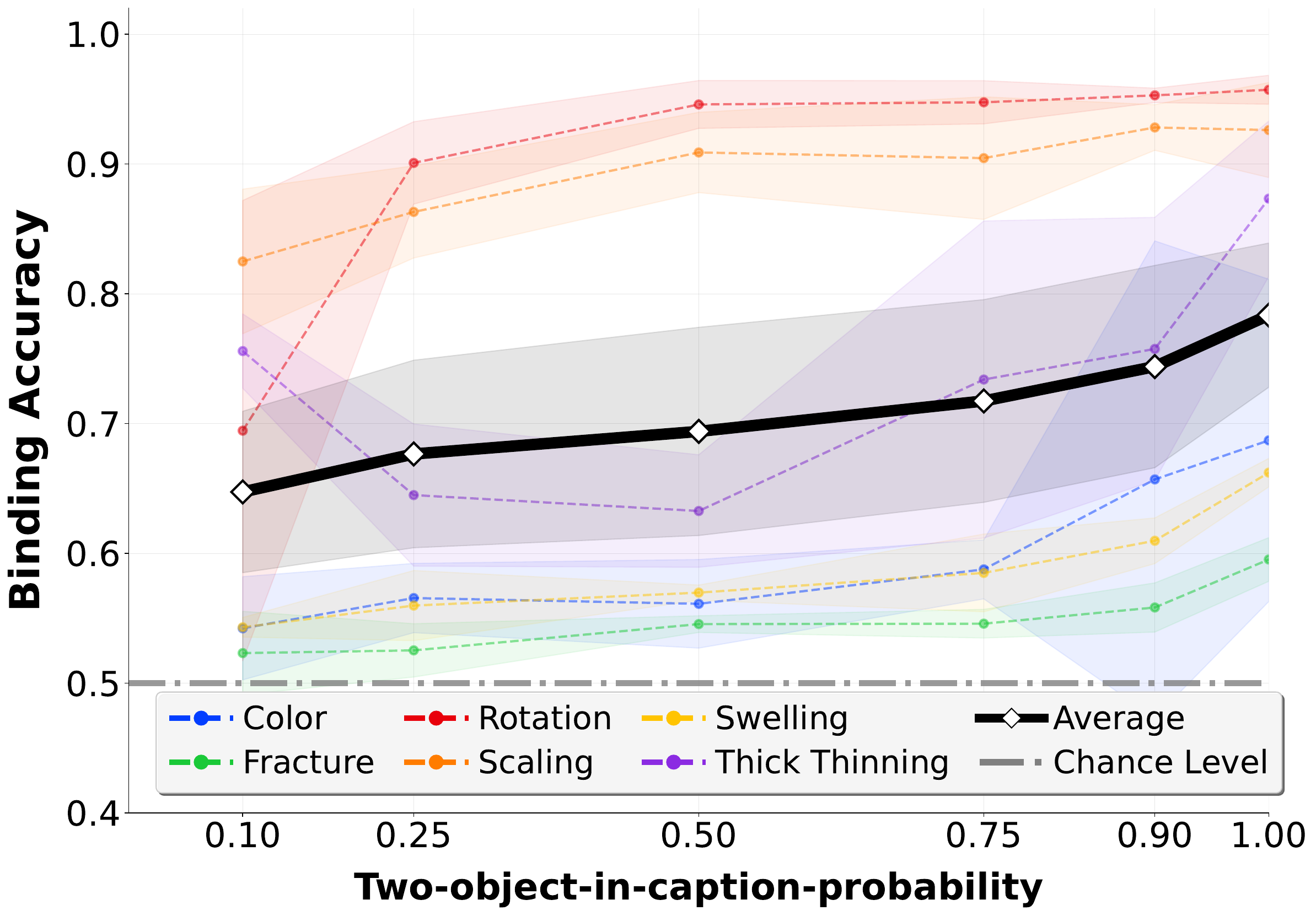}
        \caption{\textbf{\multiObjRatioCap}}
        \label{fig:multi-object_caption_ratio}
    \end{subfigure}    
    \caption{\textbf{Influence of (a) \multiObjRatioImg and (b) \multiObjRatioCap on binding accuracy on MADMAN.}
    Having more images and captions containing multiple objects improves binding.}
    \label{fig:multi-object_ratios}
\end{figure}
To answer this we vary \multiObjRatioImg.
As can be seen in \cref{fig:multi-object_image_ratio}, object-attribute binding increases steadily for all attributes.

\subsubsection{What's the influence of the ratio of two-object captions on binding?}
Similarly, the number of objects mentioned in the caption likely influences binding.
\cref{fig:multi-object_caption_ratio} reveals that a small ratio of captions containing two objects are already sufficient to learn binding for a few attributes. Increasing the number of captions containing two objects leads to consistent improvements of binding.

\subsubsection{How does the \textit{number of attributes per object in the caption} influence binding?}
\begin{figure}[t]
    \centering
    \begin{subfigure}[b]{0.49\textwidth}
        \includegraphics[width=1\linewidth]{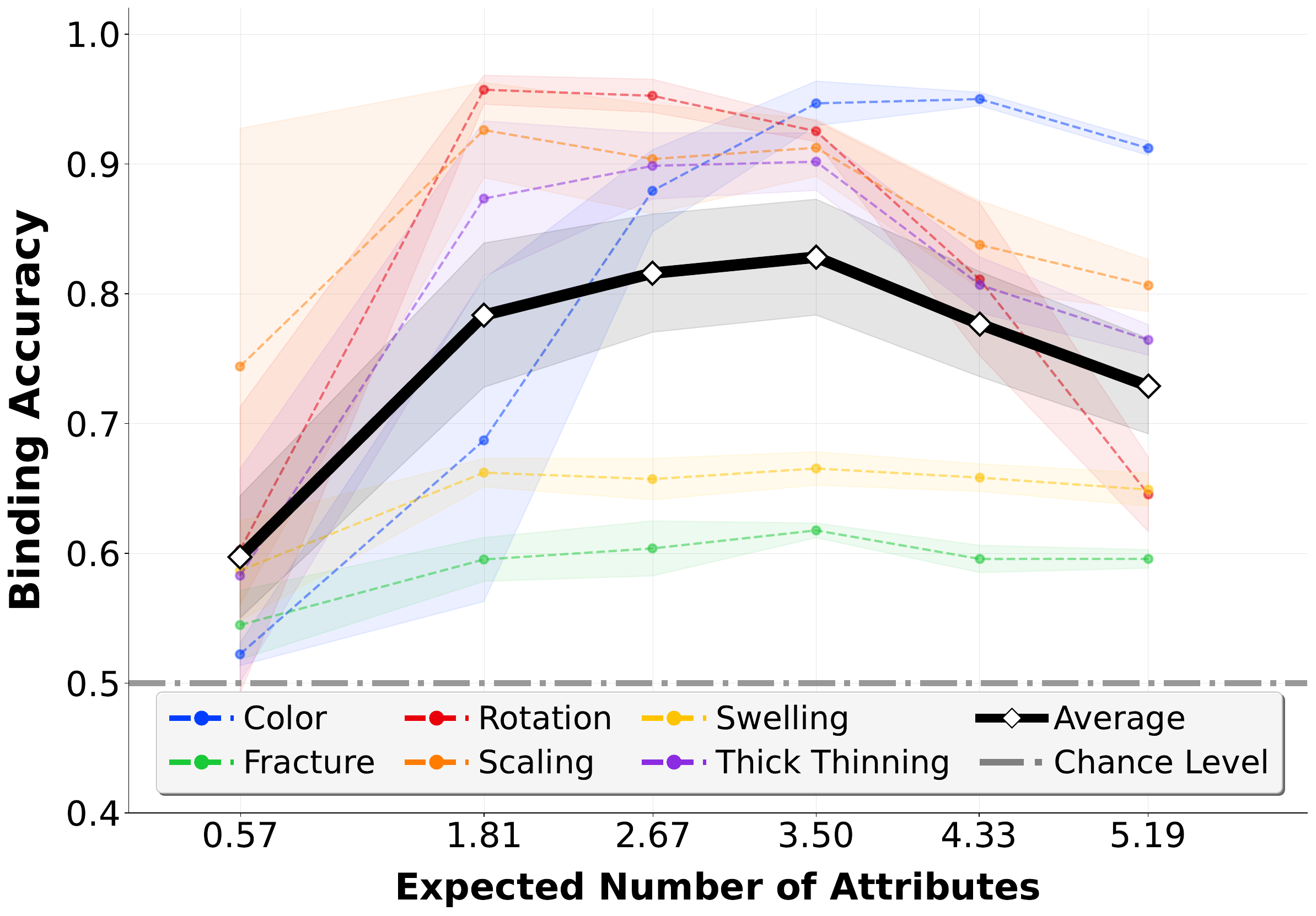}
        \caption{\textbf{\attrsInCap}}
        \label{fig:nr_attr_in_cap}
    \end{subfigure}    
    \begin{subfigure}[b]{0.49\textwidth}    
        \includegraphics[width=1\linewidth]{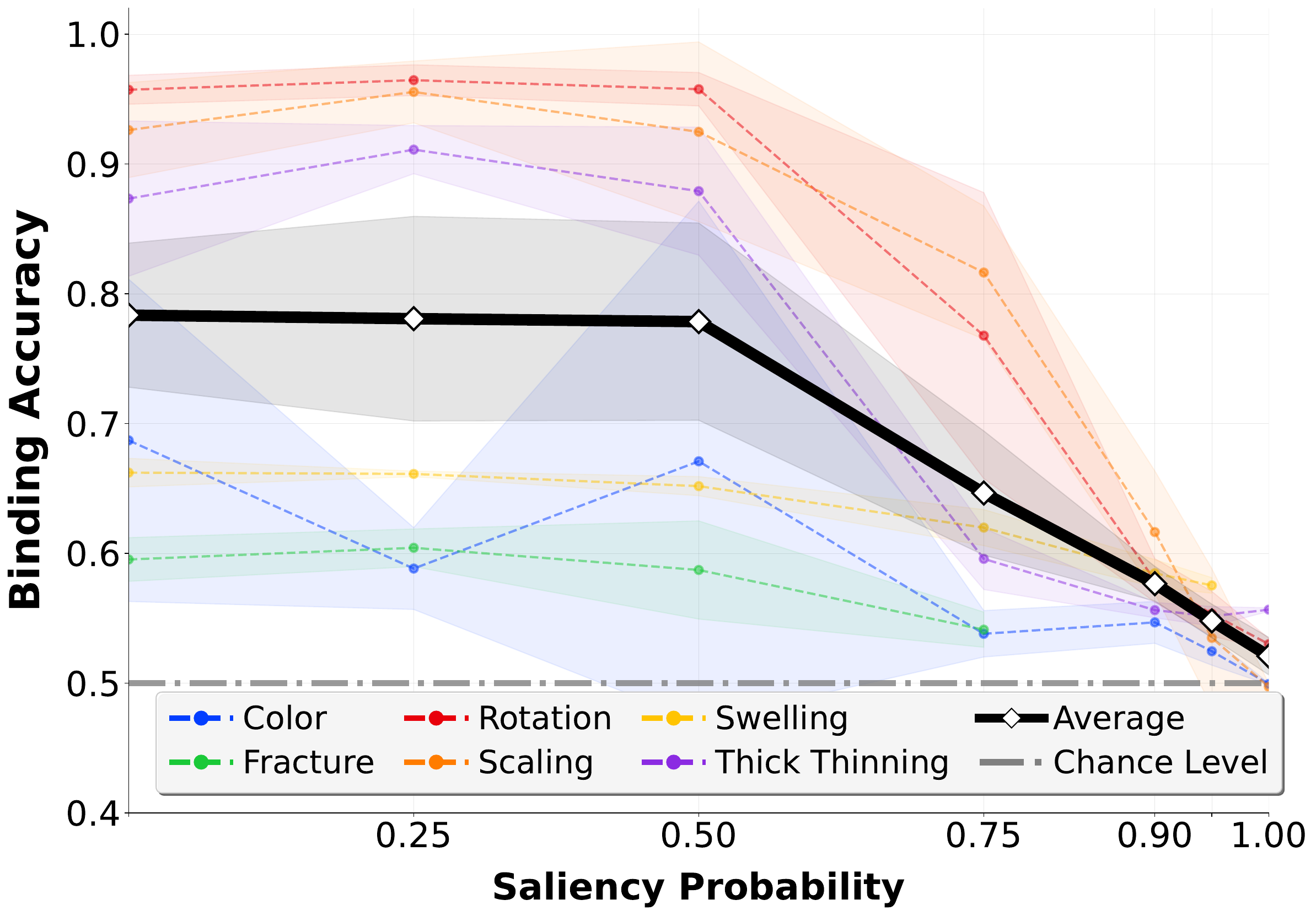}
        \caption{\textbf{\saliencyBias}}
        \label{fig:saliency}
    \end{subfigure}    
    \caption{
    \textbf{Influence of (a) \attrsInCap and (b) \saliencyBias on binding accuracy on MADMAN.}
    \textbf{(a)} Too few and too many \attrsInCap hamper the binding accuracy. 
    For large \attrsInCap binding is not necessary anymore, as BOW is sufficient to discriminate images due to combinatorial explosion. \textbf{(b)} Saliency bias hinders binding, especially at higher values, which is the case for natural data. This is likely because it provides a shortcut for the model to minimize contrastive loss by simply representing the salient object(s) without binding them.}
    \label{fig:num-attrs-saliency}
\end{figure}
As can be seen in \cref{fig:nr_attr_in_cap}, the \attrsInCap have a strong influence on binding. 
Interestingly, for \attrsInCap larger than $3.5$ the binding accuracy starts to decrease again. 
We suspect that this is due to the larger discriminativeness of a BOW representation with many concepts due to the combinatorial explosion:
If enough attributes are present, a BOW representation is sufficient to discriminate between all samples in a batch (compare \cref{fig:data-setup-poor}) and thus the model does not need to learn binding.
In the other extreme, low \attrsInCap prevents binding from being learned, as cases with two attributes where binding is necessary, are too rare.

\subsubsection{What's the influence of the saliency bias of captioners on binding?}
Humans typically agree on which object(s) are the most important (or salient) in an image \cite{salient_object_analysis,saliency_stands_out}. 
Saliency might hamper binding as it provides shortcut-opportunities to minimize the loss, e.g., by representing primarily the attributes of the salient object without binding them.\footnote{Note that the text encoder also has information on which object is likely the salient one, as the salient object is always mentioned first in the caption.}
Indeed we find that already small levels of saliency have a negative effect, while levels above 0.75 are detrimental (\cref{fig:saliency}). 

But can we really expect more than 75\% of image-caption pairs to express saliency bias?
We expect this to be the case as humans tend to take pictures of objects instead of random images and tend to describe the salient objects with higher probability.
To verify this, we use our manually labeled images of CC12M.
Indeed, we identified at least one salient object for more than 90\% of the images (see Appendix, \cref{realistic-value-procedure}).
In our experiments we observe very low binding for these saliency levels (\cref{fig:saliency}), suggesting that saliency is likely one of the main limiting factors for CLIP models to learn binding.
We even find that the influence of other factors diminishes when saliency is at a natural level (see \cref{tab:comparison_realistic_sub}). Thus saliency might be a key limiting factor also on real data.

\subsubsection{\textit{Color} often behaves differently.}
Interestingly, we observe that the attribute \textit{color} binds poorly for most values of \multiObjRatioImg (\cref{fig:multi-object_image_ratio}) and \multiObjRatioCap (\cref{fig:multi-object_caption_ratio}), despite being easy to recognize (see Appendix, \cref{recognition_in_distribution}).
But why does it bind so poorly?
Note that the color attribute can take 7 different values, while all other attributes have fewer possible values. As a result, color has most discriminative power among the attributes, e.g., knowing that there is a \textit{green 6} in the image distinguishes it from more images than knowing there is a \textit{small 6}.
Thus, if color(s) and object(s) are in the caption, representing these two factors is often sufficient to distinguish this image/caption from all other captions/images in a batch, making binding unnecessary.
Only when colors of more than one object are frequently in the caption, learning the color-object binding reduces the loss (relatively) enough s.t.~the model eventually learns binding.

\noindent\textbf{The ideal data setting.}
Finally, we use the previous experiments to select the best data setup for each of the properties. We end up with: no saliency bias, $p(\text{multi-obj-img})=1$, $p(\text{multi-obj-cap})=1$ and expected number of \attrsInCap= 3.5.
As shown in \cref{tab:teaser_tab}, combining the optimal setups for all our data properties indeed leads to high binding and combining the best setups for each parameter leads to best binding performance (see Appendix, \cref{individual_ideal_setup_performance}).
\cref{tab:teaser_tab} reveals that ideal data alone leads to high object-attribute binding for 4 out of 6 attributes.
Using ideal data, a large model and a large batch size results in almost perfect binding for all attributes.

\begin{figure}[ht]
    \centering
    \begin{subfigure}[b]{0.99\textwidth}    
        \includegraphics[width=1\linewidth]{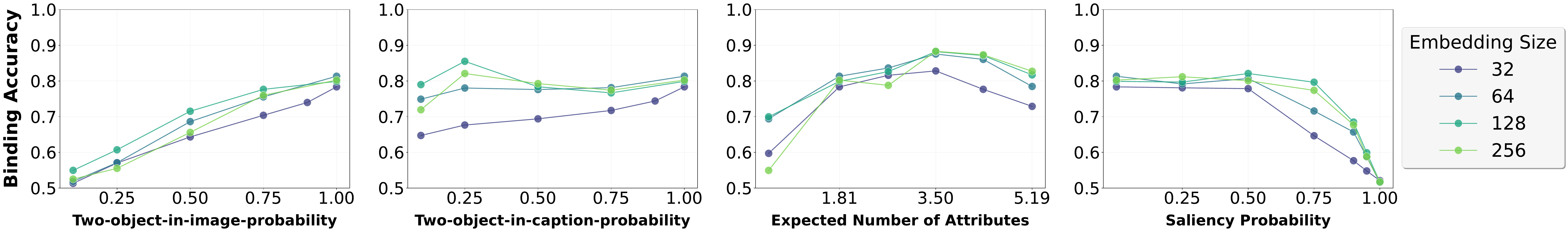}
        \vspace{-1.25\baselineskip}
        \caption{Validating embedding size choice}
         \label{fig:validating_setup_embed}      
    \end{subfigure}  

    \vspace{0.5\baselineskip}

    \begin{subfigure}[b]{0.99\textwidth}    
        \includegraphics[width=1\linewidth]{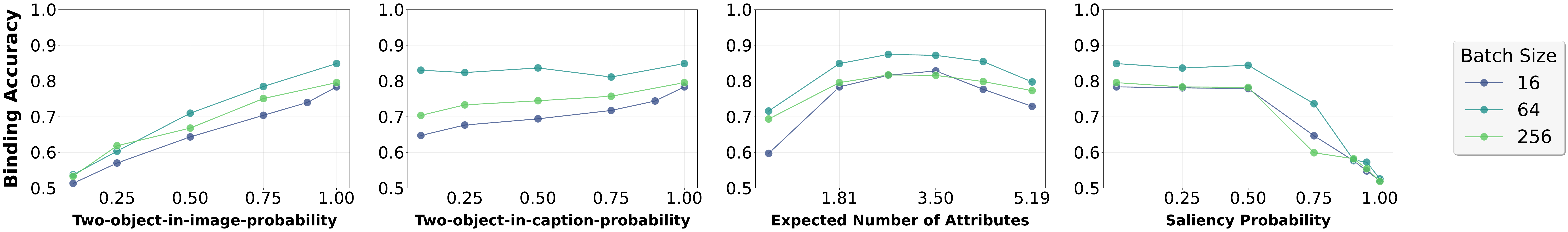}
        \vspace{-1.25\baselineskip}
        \caption{Validating batch size choice using an embedding size of 32}
         \label{fig:validating_setup_batchSizeSmall}      
    \end{subfigure}    

    \vspace{0.5\baselineskip}
    
     \begin{subfigure}[b]{0.99\textwidth} 
        \includegraphics[width=1\linewidth]{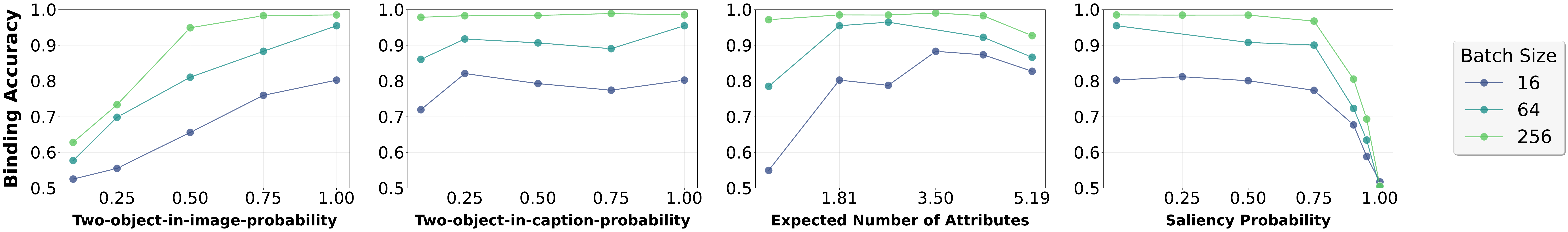}
        \vspace{-1.25\baselineskip}
        \caption{Validating batch size using an embedding size of 256}
         \label{fig:validating_setup_batchSizeLarge}      
    \end{subfigure}
     \vspace{-0.75\baselineskip}
    \caption{\textbf{Experimental results are independent of most important hyper-parameters of the experimental setup on MADMAN.} We test various combinations of the embedding size and batch size. We observe the same trends for all data properties as found in our main setting, showing that our results are not an artifact of a specific experimental setup.
    }
    \label{fig:validing_the_setup}
\end{figure}

\subsection{Validating our Experimental Setup}
\label{validating_the_setup}
The results presented above could be an artifact of an unlucky choice of the batch size or model size.
To rule out that our findings are just an artifact of poor choices for the batch size and model size, 
we repeat the experiments and vary the batch size and the embedding size of the models.
As detailed above, the batch size directly controls the number of hard negatives in the loss, thus testing different batch sizes can be interpreted as changing the hard negative ratio in a batch. 
By changing the embedding size we test whether the model size limited binding in our experiments.
\cref{fig:validating_setup_embed} shows that our results are not an artifact of the embedding size. We observe small differences, however, the trends remain the same.
Similarly, \cref{fig:validating_setup_batchSizeSmall,fig:validating_setup_batchSizeLarge} reveal that our findings are independent of the batch size.
Changing the batch size leads to small differences, but the trends remain consistent.
Combined, these results validate our experimental setting and findings as not being an artifact of the base setup.
The only exception of this is the \multiObjRatioCap, where the effect partially diminishes with increasing embedding size and increasing batch size (\cref{fig:validating_setup_embed,fig:validating_setup_batchSizeSmall,fig:validating_setup_batchSizeLarge}).

\subsection{We find the Same Relationships Out-of-distribution} \label{ood-results}

\begin{figure}[ht]
    \centering
    \begin{subfigure}[b]{0.49\textwidth}
        \includegraphics[width=1\linewidth]{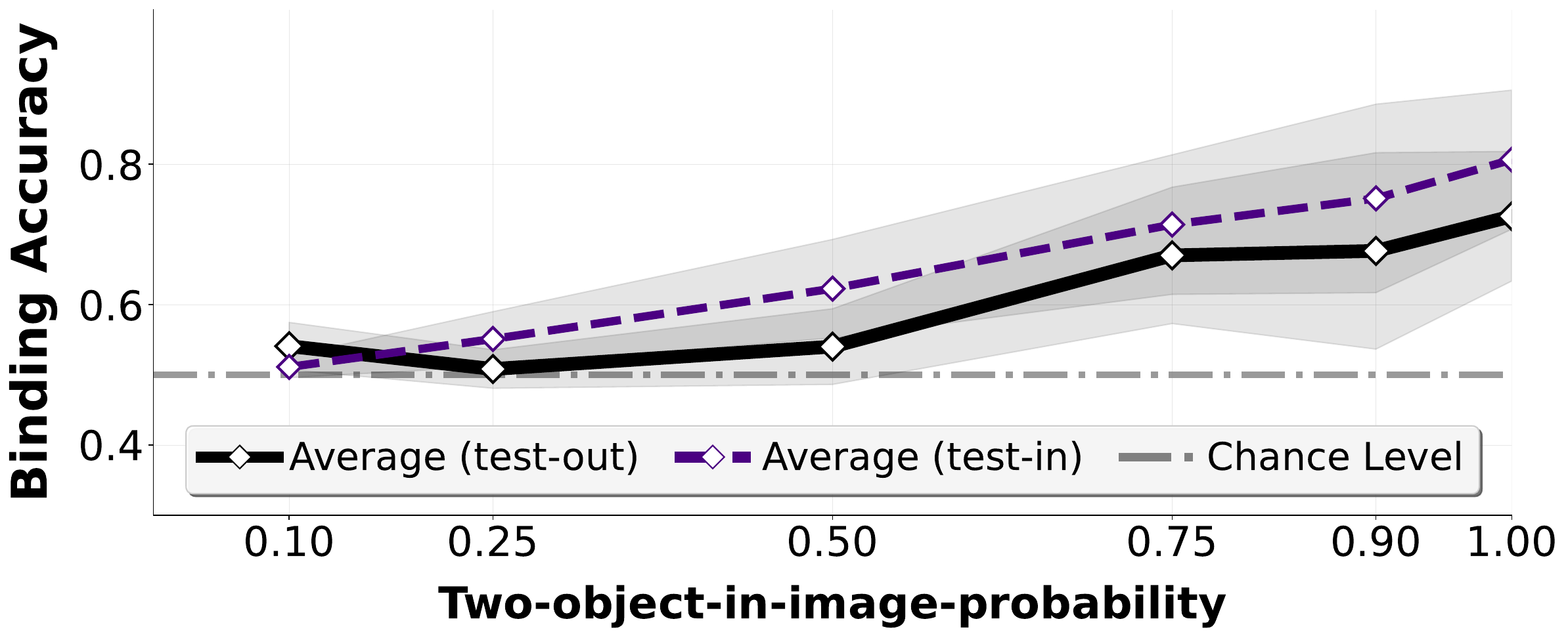}
        \vspace{-1.5\baselineskip}
        \caption{\textbf{\multiObjRatioImg}}
        \label{fig:test-out-multi-object_image_ratio}
    \end{subfigure}
    \begin{subfigure}[b]{0.49\textwidth}
        \includegraphics[width=1\linewidth]{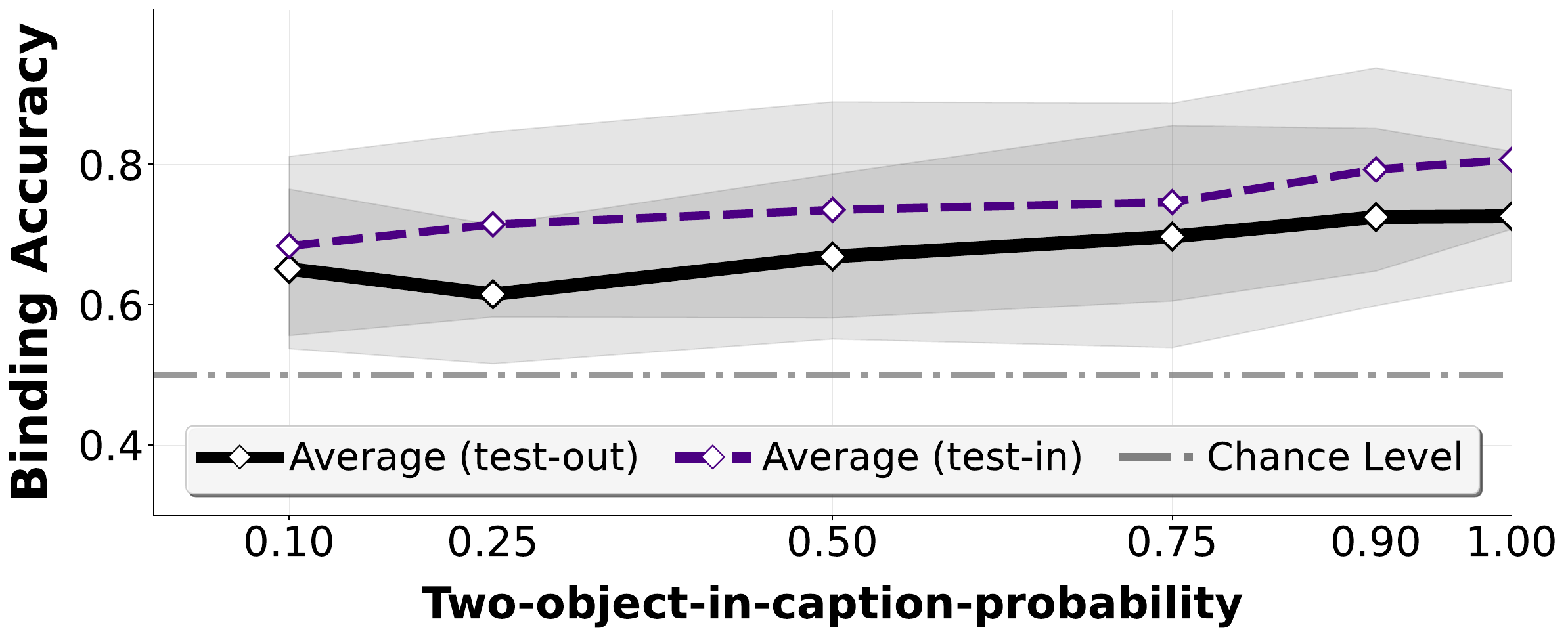}
        \vspace{-1.5\baselineskip}
        \caption{\textbf{\multiObjRatioCap}}
        \label{fig:test-out-multi-object_caption_ratio}
    \end{subfigure}
    
    \vspace{0.5\baselineskip} 

    \begin{subfigure}[b]{0.49\textwidth}
        \includegraphics[width=1\linewidth]{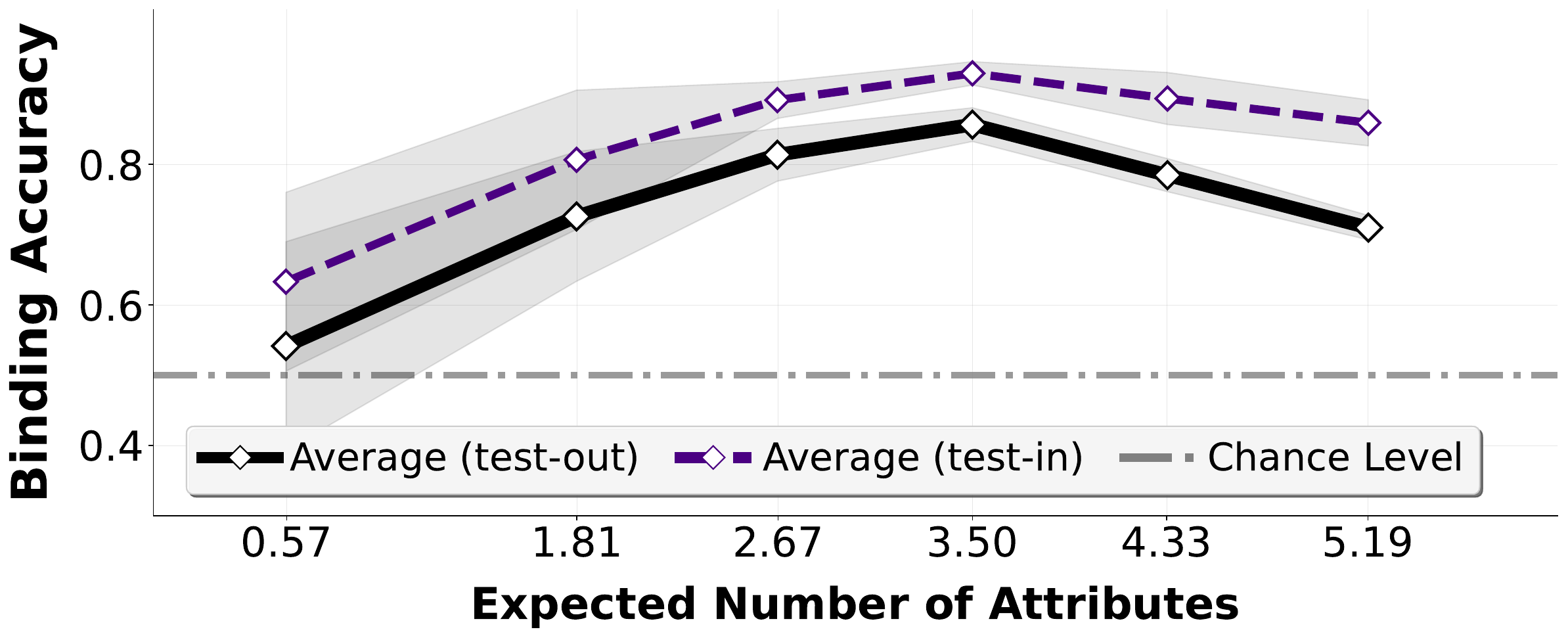}
        \vspace{-1.5\baselineskip}
        \caption{\textbf{\attrsInCap}}
        \label{fig:test-out-nr_attr_in_cap}
    \end{subfigure}
    \begin{subfigure}[b]{0.49\textwidth}
        \includegraphics[width=1\linewidth]{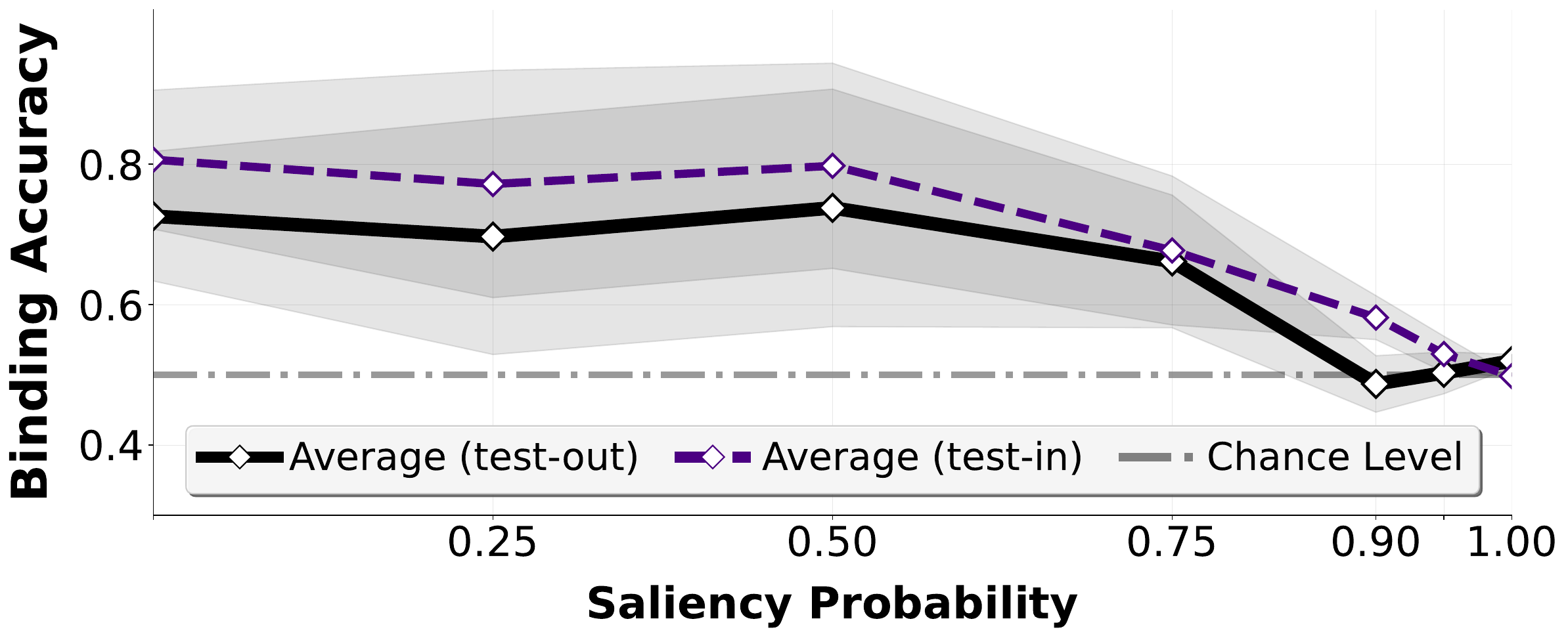}
        \vspace{-1.5\baselineskip}
        \caption{\textbf{\saliencyBias}}        
        \label{fig:test-out-saliency}
    \end{subfigure}
    
    \caption{\textbf{Impact of the data properties on the binding for\textbf{ out-of-distribution} combinations on MADMAN-OOD}. It follows similar trends as in-distribution (denoted by the dashed line).}
    \label{fig:out-of-distribution-results}
\end{figure}

CLIP models have been shown to learn object-attribute compositional representations \cite{idealWords}, meaning that the concept ``yellow submarine'' is encoded roughly by the sum of the ``yellow'' vector and the ``submarine'' vector.
However, in a synthetic setting with few factors of variation a 
model can just treat each object- attribute combination as an individual class, instead of learning such a compositional representation. 
If our models learn non-compositional representations we can not expect our results to transfer to large scale CLIP training.
To test this, we test our models on our OOD test set with unseen object-attribute combinations. 
A compositional representation can generalize to unseen object-attribute combinations, while a non-compositional representation can not.
We also find the same trends for OOD (see \cref{fig:out-of-distribution-results}). Thus, our model learns a similar compositional representation as CLIP models trained on real data, which confirms the validity of our study.

\subsection{Are Hard Negatives Enough to Solve the Binding Problem?}

\begin{table}[t!]
\centering
\caption{\textbf{Attribute binding Performance of NegCLIP} with realistic data. The base CLIP when provided with an ideal data setup learns the binding much better, highlighting the importance of getting the right data properties.}
\begin{tabular}{lcccccc}
\hline
\textbf{Model} & \textbf{Color} & \textbf{Scaling} & \textbf{Fracture} & \textbf{Rotation} & \textbf{Swelling} & \textbf{Thickness} \\
\hline
CLIP                  & 50.47 & 54.47 & 52.34 & 50.86 & 53.26 & 50.05 \\
CLIP (+Ideal data) & 94.66 & 91.28 & 61.93 & 92.44 & 66.43 & 90.02 \\
NegCLIP (Text)         & 77.10 & 75.65 & 54.84 & 74.83 & 56.44 & 62.39 \\
NegCLIP (Text+Image)   & 72.84 & 76.85 & 56.14 & 86.74 & 58.18 & 63.93 \\
\hline
\end{tabular}
\label{tab:negclip_transposed}
\end{table}
Previous works tried to resolve binding by changing the training process, for instance by training on hard negatives \cite{clip_bow}.
The previous sections show that data properties of the training data have a large influence on the binding performance of CLIP models.
Naturally, the question arises whether a hard negative training regime, as proposed for NegCLIP \cite{clip_bow}, can be used to resolve the problems induced by the data properties.
To this end, we explore how well NegCLIP performs on the variant of MADMAN following realistic data properties. 
As \cref{tab:negclip_transposed} reveals, NegCLIP indeed improves binding over CLIP, but does not get close to CLIP trained in the ideal data setup. Thus, NegCLIP alone is not enough to resolve the issues caused by the data properties and thus is unlikely to resolve the issue on real data. 
Our results indicate that better training data might be necessary to resolve the binding problem of CLIP.

\section{Conclusion}

In this work, we showed that CLIP struggles to learn object-attribute binding even when training under favorable conditions.
Via a rigorous analysis using our synthetically generated dataset we showed that specific data properties common in web-scraped image-caption datasets are limiting CLIP's object-attribute binding. 
In particular, we found that having too few images showing, and too few captions describing multiple objects limits binding. More importantly, we showed that too few \textit{and} too many attributes mentioned in the caption per object hurt binding.
Most interestingly, we found that the saliency bias plays a critical role and, at levels common on real data, has a detrimental effect on the binding performance.
Furthermore, we found that when the saliency bias is too high, the influence of the other properties diminishes.
We verified our experiments thoroughly, by changing model parameters and testing in an OOD setup, where we found the same trends. 
In summary, our controlled experiments on synthetic data provide evidence that the data properties common for natural image-caption datasets have a negative influence on CLIP's binding performance. 
To obtain high binding performance of CLIP models on natural data it might be necessary to change some of the data properties of the training data.

\noindent\textbf{Limitations.}
This work shows the relationship of the aforementioned data properties to binding accuracy of CLIP models; however, our studies are limited to a synthetic setup.
We expect to find the same relationship on real data but obtaining the required information to perform these experiments on real data is prohibitively expensive. 

\noindent\textbf{Future work} should use the insights provided by this work to guide filtering or re-captioning approaches.
A promising re-captioning approach is to occasionally drop the salient object from the caption.
Another promising direction is to explore the optimal number of attributes per object in the caption on real data.

\section*{Acknowledgments}
We would like to thank Simon Schrodi for helpful comments on the draft.
Funded by the German Federal Ministry for Economic Affairs and Energy within the project “NXT GEN AI METHODS" (19A23014R), by the German Research Foundation (DFG) - 417962828, 539134284 and – SFB 1597 – 499552394. 

\bibliographystyle{splncs04}
\bibliography{bibliography}

\begin{thebibliography}{10}
\providecommand{\url}[1]{\texttt{#1}}
\providecommand{\urlprefix}{URL }
\providecommand{\doi}[1]{https://doi.org/#1}

\bibitem{oc-clip}
Assouel, R., Astolfi, P., Bordes, F., Drozdzal, M., Romero-Soriano, A.: Object-centric binding in contrastive language-image pretraining. arXiv preprint arXiv:2502.14113  (2025)

\bibitem{beyer2024paligemma}
Beyer, L., Steiner, A., Pinto, A.S., Kolesnikov, A., Wang, X., Salz, D., Neumann, M., Alabdulmohsin, I., Tschannen, M., Bugliarello, E., et~al.: Paligemma: A versatile 3b vlm for transfer. arXiv preprint arXiv:2407.07726  (2024)

\bibitem{pugspar}
Bordes, F., Shekhar, S., Ibrahim, M., Bouchacourt, D., Vincent, P., Morcos, A.: Pug: Photorealistic and semantically controllable synthetic data for representation learning. Advances in Neural Information Processing Systems  \textbf{36},  45020--45054 (2023)

\bibitem{salient_object_analysis}
Borji, A., Sihite, D.N., Itti, L.: Quantitative analysis of human-model agreement in visual saliency modeling: A comparative study. IEEE Transactions on Image Processing  \textbf{22}(1),  55--69 (2012)

\bibitem{saliency_stands_out}
Borji, A., Sihite, D.N., Itti, L.: What stands out in a scene? a study of human explicit saliency judgment. Vision research  \textbf{91},  62--77 (2013)

\bibitem{castro2019morpho}
Castro, D.C., Tan, J., Kainz, B., Konukoglu, E., Glocker, B.: Morpho-mnist: Quantitative assessment and diagnostics for representation learning. Journal of Machine Learning Research  \textbf{20}(178),  1--29 (2019)

\bibitem{changpinyo2021cc12m}
Changpinyo, S., Sharma, P., Ding, N., Soricut, R.: {Conceptual 12M}: Pushing web-scale image-text pre-training to recognize long-tail visual concepts. In: CVPR (2021)

\bibitem{chen2023pali}
Chen, X., Wang, X., Changpinyo, S., Piergiovanni, A., Padlewski, P., Salz, D., Goodman, S., Grycner, A., Mustafa, B., Beyer, L., Kolesnikov, A., Puigcerver, J., Ding, N., Rong, K., Akbari, H., Mishra, G., Xue, L., Thapliyal, A.V., Bradbury, J., Kuo, W., Seyedhosseini, M., Jia, C., Ayan, B.K., Ruiz, C.R., Steiner, A.P., Angelova, A., Zhai, X., Houlsby, N., Soricut, R.: Pa{LI}: A jointly-scaled multilingual language-image model. In: The Eleventh International Conference on Learning Representations (2023), \url{https://openreview.net/forum?id=mWVoBz4W0u}

\bibitem{openclip}
Cherti, M., Beaumont, R., Wightman, R., Wortsman, M., Ilharco, G., Gordon, C., Schuhmann, C., Schmidt, L., Jitsev, J.: Reproducible scaling laws for contrastive language-image learning. In: Proceedings of the IEEE/CVF Conference on Computer Vision and Pattern Recognition. pp. 2818--2829 (2023)

\bibitem{doveh2023dense}
Doveh, S., Arbelle, A., Harary, S., Herzig, R., Kim, D., Cascante-Bonilla, P., Alfassy, A., Panda, R., Giryes, R., Feris, R., et~al.: Dense and aligned captions (dac) promote compositional reasoning in vl models. Advances in Neural Information Processing Systems  \textbf{36},  76137--76150 (2023)

\bibitem{doveh2023teaching}
Doveh, S., Arbelle, A., Harary, S., Schwartz, E., Herzig, R., Giryes, R., Feris, R., Panda, R., Ullman, S., Karlinsky, L.: Teaching structured vision \& language concepts to vision \& language models. In: Proceedings of the IEEE/CVF Conference on Computer Vision and Pattern Recognition. pp. 2657--2668 (2023)

\bibitem{sugarcrepe++}
Dumpala, S.H., Jaiswal, A., Shama~Sastry, C., Milios, E., Oore, S., Sajjad, H.: Sugarcrepe++ dataset: Vision-language model sensitivity to semantic and lexical alterations. Advances in Neural Information Processing Systems  \textbf{37},  17972--18018 (2024)

\bibitem{hsieh2023sugarcrepe}
Hsieh, C.Y., Zhang, J., Ma, Z., Kembhavi, A., Krishna, R.: Sugarcrepe: Fixing hackable benchmarks for vision-language compositionality. Advances in neural information processing systems  \textbf{36},  31096--31116 (2023)

\bibitem{clevr_dataset2017}
Johnson, J., Hariharan, B., Van Der~Maaten, L., Fei-Fei, L., Zitnick, C.L., Girshick, R.: {CLEVR}: {A} {Diagnostic} {Dataset} for {Compositional} {Language} and {Elementary} {Visual} {Reasoning}. In: 2017 {IEEE} {Conference} on {Computer} {Vision} and {Pattern} {Recognition} ({CVPR}). pp. 1988--1997. IEEE, Honolulu, HI (Jul 2017). \doi{10.1109/CVPR.2017.215}, \url{https://ieeexplore.ieee.org/document/8099698/}

\bibitem{clip_unimodal_bow}
Koishigarina, D., Uselis, A., Oh, S.J.: Clip behaves like a bag-of-words model cross-modally but not uni-modally. arXiv preprint arXiv:2502.03566  (2025)

\bibitem{lecun2010mnist}
LeCun, Y., Cortes, C., Burges, C.: Mnist handwritten digit database. ATT Labs [Online]. Available: http://yann.lecun.com/exdb/mnist  \textbf{2} (2010)

\bibitem{clip_bind_concept}
Lewis, M., Nayak, N., Yu, P., Merullo, J., Yu, Q., Bach, S., Pavlick, E.: Does {CLIP} bind concepts? probing compositionality in large image models. In: Graham, Y., Purver, M. (eds.) Findings of the Association for Computational Linguistics: EACL 2024. pp. 1487--1500. Association for Computational Linguistics, St. Julian{'}s, Malta (Mar 2024), \url{https://aclanthology.org/2024.findings-eacl.101/}

\bibitem{declip}
Li, Y., Liang, F., Zhao, L., Cui, Y., Ouyang, W., Shao, J., Yu, F., Yan, J.: Supervision exists everywhere: A data efficient contrastive language-image pre-training paradigm. In: International Conference on Learning Representations (2022), \url{https://openreview.net/forum?id=zq1iJkNk3uN}

\bibitem{crepe2023}
Ma, Z., Hong, J., Gul, M.O., Gandhi, M., Gao, I., Krishna, R.: Crepe: Can vision-language foundation models reason compositionally? In: Proceedings of the IEEE/CVF Conference on Computer Vision and Pattern Recognition. pp. 10910--10921 (2023)

\bibitem{nichol2021glide}
Nichol, A., Dhariwal, P., Ramesh, A., Shyam, P., Mishkin, P., McGrew, B., Sutskever, I., Chen, M.: Glide: Towards photorealistic image generation and editing with text-guided diffusion models. arXiv preprint arXiv:2112.10741  (2021)

\bibitem{patel2024tripletclip}
Patel, M., Kusumba, N.S.A., Cheng, S., Kim, C., Gokhale, T., Baral, C., et~al.: Tripletclip: Improving compositional reasoning of clip via synthetic vision-language negatives. Advances in neural information processing systems  \textbf{37},  32731--32760 (2024)

\bibitem{clip}
Radford, A., Kim, J.W., Hallacy, C., Ramesh, A., Goh, G., Agarwal, S., Sastry, G., Askell, A., Mishkin, P., Clark, J., et~al.: Learning transferable visual models from natural language supervision. In: International conference on machine learning. pp. 8748--8763. PmLR (2021)

\bibitem{ramesh2022hierarchical}
Ramesh, A., Dhariwal, P., Nichol, A., Chu, C., Chen, M.: Hierarchical text-conditional image generation with clip latents. arXiv preprint arXiv:2204.06125  \textbf{1}(2), ~3 (2022)

\bibitem{rombach2022high}
Rombach, R., Blattmann, A., Lorenz, D., Esser, P., Ommer, B.: High-resolution image synthesis with latent diffusion models. In: Proceedings of the IEEE/CVF conference on computer vision and pattern recognition. pp. 10684--10695 (2022)

\bibitem{twoEffects}
Schrodi, S., Hoffmann, D.T., Argus, M., Fischer, V., Brox, T.: Two effects, one trigger: On the modality gap, object bias, and information imbalance in contrastive vision-language model. In: International Conference on Learning Representations (ICLR) (2025)

\bibitem{lemonsPurple}
Tang, Y., Yamada, Y., Zhang, Y., Yildirim, I.: When are lemons purple? the concept association bias of vision-language models. In: Proceedings of the 2023 Conference on Empirical Methods in Natural Language Processing. pp. 14333--14348 (2023)

\bibitem{idealWords}
Trager, M., Perera, P., Zancato, L., Achille, A., Bhatia, P., Soatto, S.: Linear spaces of meanings: compositional structures in vision-language models. In: Proceedings of the IEEE/CVF International Conference on Computer Vision. pp. 15395--15404 (2023)

\bibitem{MetaCLIP}
Xu, H., Xie, S., Tan, X., Huang, P.Y., Howes, R., Sharma, V., Li, S.W., Ghosh, G., Zettlemoyer, L., Feichtenhofer, C.: Demystifying {CLIP} data. In: The Twelfth International Conference on Learning Representations (2024), \url{https://openreview.net/forum?id=5BCFlnfE1g}

\bibitem{clip_bow}
Yuksekgonul, M., Bianchi, F., Kalluri, P., Jurafsky, D., Zou, J.: When and why vision-language models behave like bags-of-words, and what to do about it? In: International Conference on Learning Representations (2023), \url{https://openreview.net/forum?id=KRLUvxh8uaX}

\bibitem{siglip}
Zhai, X., Mustafa, B., Kolesnikov, A., Beyer, L.: Sigmoid loss for language image pre-training. In: Proceedings of the IEEE/CVF international conference on computer vision. pp. 11975--11986 (2023)

\bibitem{zhong2022regionclip}
Zhong, Y., Yang, J., Zhang, P., Li, C., Codella, N., Li, L.H., Zhou, L., Dai, X., Yuan, L., Li, Y., et~al.: Regionclip: Region-based language-image pretraining. In: Proceedings of the IEEE/CVF Conference on Computer Vision and Pattern Recognition. pp. 16793--16803 (2022)

\end{thebibliography}
\clearpage
\section*{Supplemental Material}

\subsection{Dataset Creation and Model Setup}
\label{creating_madman}

We introduce MADMAN, a synthetic dataset designed to systematically investigate how dataset properties affect object-attribute binding. The dataset enables precise control over four key aspects: the number of objects per image, the completeness of object descriptions in captions, the density of attribute descriptions per object, and saliency bias.

Each image contains multiple objects with controllable attributes, placed in a structured grid layout. The corresponding captions are generated using a template-based approach that allows explicit control over which objects and attributes are described. This design facilitates systematic experimentation by enabling independent manipulation of image content and caption completeness.

The following subsections detail the dataset construction process, including object creation, image composition, and caption generation. We also formally define the data properties that can be manipulated to study their effects on attribute binding.

\subsubsection{Creating Objects with attributes.}

Firstly, we need to create objects with attributes.
We build on the MAD dataset from Schrodi et al.~\cite{twoEffects}, which is derived from Morpho-MNIST \citep{castro2019morpho}.

MAD uses MNIST \citep{lecun2010mnist} digits as objects and applies a set of transformations on them representing different attributes: thin/thickness, swelling, fracture, scaling, and coloring. We add rotation as another transformation / attribute to have one more factor of variation. 

\begin{table}[!ht]
\centering
\caption[List of Transformations used as attributes of the digits]{Transformations applied on the digits. The transformation categories act as the attributes and the type within each category that we apply to the digit is the value of the attribute. We thus have $(3 * 2 * 2 * 2 * 3 * 7)=504$ possible combinations of attribute values.}
\label{table:attribute-values}
\begin{tabular}{|l|l|}
\hline
\textbf{Attribute} & \textbf{Values} \\
\hline
thickness & \texttt{no-thickthinning} : Identity \\
                & \texttt{thickening} : Thicken by 0.7 of its thickness.\\
                & \texttt{thinning} : Thin a digit by 0.7 of its thickness. \\
\hline
swelling & \texttt{no-swelling} : Identity \\
         & \texttt{swelling} : Create a local swelling at a random location.\\
\hline
fracture & \texttt{no-fracture} : Identity \\
         & \texttt{fracture} : Add fractures to random locations on the digit. \\
\hline
scaling & \texttt{large} : Identity \\
        & \texttt{small} : Rescale to 75\% of its size.\\ 
\hline
rotation & \texttt{no-rotation} : Identity \\
         & \texttt{rotate-p36} : Rotate +36$^{\circ}$ i.e. anti-clockwise.\\
         & \texttt{rotate-n36} : Rotate -36$^{\circ}$ i.e. clockwise.\\
\hline
color & One of \{ \texttt{gray}, \texttt{red}, \texttt{green}, \texttt{blue}, \texttt{cyan}, \texttt{magenta}, \texttt{yellow} \} \\
\hline
\end{tabular}
\end{table}

\subsubsection{Creating the (multi-object) images.}

To create a multi-object setup, we create a 3x3 grid, consisting of 9 cells of 28x28 pixels each, and fill two of them with samples from MAD above to create a two-object image of size 96x96.

The generation process is as follows:

\begin{enumerate}
    \item Select $o_{i}$=(1 or 2) objects to add to the image. We vary this using a parameter $p(\text{two-obj-img})$ that determines whether the second object is included or not.
    \item Randomly select the n cell locations within the 3x3 grid and place the objects making sure they do not overlap, i.e. they are not inserted in the same cells.
\end{enumerate}

\subsubsection{Creating the captions.}

When creating the corresponding captions for the image, we want to be able to vary how much of the information in the image is captured in the caption. In this synthetic setup we can have a very high degree of caption completeness in terms of information conveyed about the picture.

We generate captions by treating attributes and digit classes as individual tokens. For each object, we create a noun phrase with the attributes appearing before the digit class. Next, we combine the noun phrases using \texttt{and} as the conjunction token. All tokens in the caption are separated by the \texttt{space} character.
Note: as listed in \cref{table:attribute-values}, we have 19 unique attribute values for the six attributes. With the ten object classes and four special tokens (\texttt{and}, \texttt{<pad>}, \texttt{<start>}, \texttt{<end>}), our vocabulary size is 33 in total.
The generation process is as follows:

\begin{enumerate}
    \item Select $o_t$=(1 or 2) objects to add to the caption. We vary this by a parameter $p(\text{two-obj-cap} | \text{two-obj-img}=\text{True})$ that determines whether the second object is included or not. Note: if there is only one object in the image, we can take only that one object in the caption.
    \item Create a caption for each object using the attributes and the object class as words.
    \begin{itemize}
        \item We vary the number of attributes ($n_a$)\label{notation:number-of-attributes} used to describe an object by sampling it from a categorical distribution (see \cref{fig:attr-dists}).
        \item When stringing together the words, we separate them with a space in between and follow a predetermined order for the attributes with the object class coming at the end.
    \end{itemize}
    \item Combine the captions for the $o_t$ objects separated by the word \texttt{and}.
\end{enumerate}

\subsubsection{Dataset Properties of interest.}

With the above data generation process, we can vary the data properties of interest. Notably, we can also vary image and captions independently to create an imbalance in the information contents. 

We explicitly list and define the data properties of interest below:

\textbf{How often are there multiple objects in the image?}

Real-world images can contain $o_i \ge 1$ objects. We capture this in our minimal setting by having $o_i \in \{1, 2\}$ controlled by \multiObjRatioImg $p(\text{two-obj-img})$.
$$
o_i = 
\begin{cases} 
 2 & \text{with probability } p(\text{two-obj-img}), \\
 1 & \text{with probability } 1 - p(\text{two-obj-img}).
\end{cases}
$$

\textbf{How often are the objects described in the caption?}

Given an image with multiple objects (in our case, two), a caption may or may not describe all of them. We capture this aspect in our setting by having $o_T \in \{1, 2\}$ controlled by \multiObjRatioCap $p(\text{two-obj-cap})$.
$$
o_T = 
\begin{cases} 
 2 & \text{with probability } $p(\text{two-obj-cap})$, \\
 1 & \text{with probability } 1 - $p(\text{two-obj-cap})$
\end{cases}
$$

Note: this is applied conditional on the images having two objects.

\textbf{How many attributes are described for each object?}

\begin{figure}
    \centering
    \includegraphics[width=1.02\linewidth]{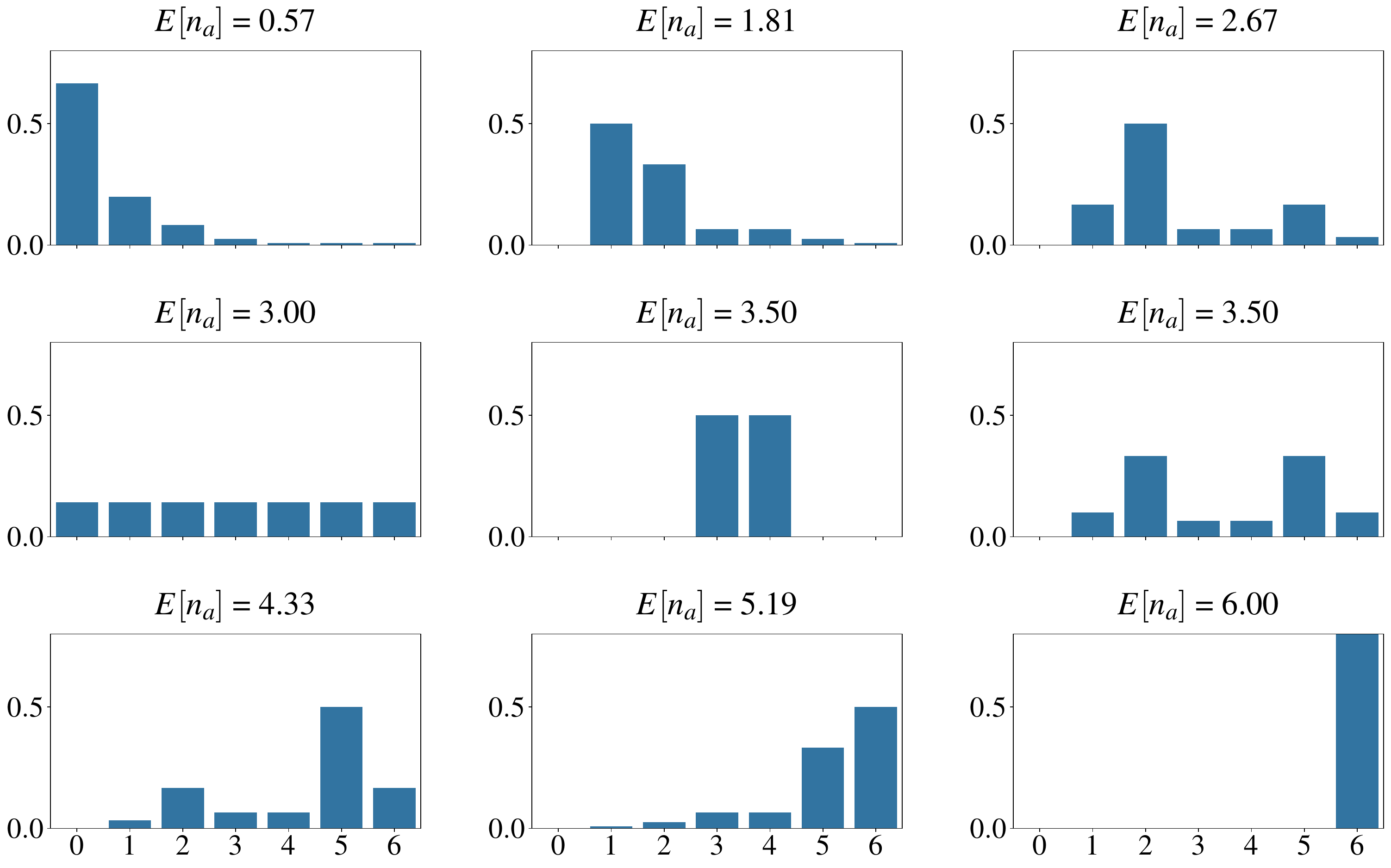}
    \caption[Distributions for number of attributes per object ($n_a$) in the captions]{Distributions for number of attributes per object ($n_a$) in the captions. We vary the distribution to cover a range of $\mathbf{E}[n_a]$. We set up the distributions such that most of them have the same probability mass for $n_a \in \{3, 4\}$ to provide a common ground for evaluation.}
    \label{fig:attr-dists}
\end{figure}

One data property we are interested in is the number of attributes associated with an object that are included in the captions. As seen in \cref{fig:attr-dists}, we define a categorical distribution for this number $n_a$ over the possible number of attributes that can be included i.e. [0, 6]. The different distributions simulate having less / more of the attributes being included per object description in the caption.

\textbf{Degree of Saliency Bias.}
Natural images frequently exhibit saliency bias --- a systematic tendency where certain objects occupy prominent positions (typically central) and receive preferential treatment in corresponding captions. This bias manifests as both spatial prominence in the image and descriptive priority in the caption, while less salient objects may be omitted entirely from textual descriptions.

\begin{figure}
    \centering
    \includegraphics[width=0.8\linewidth]{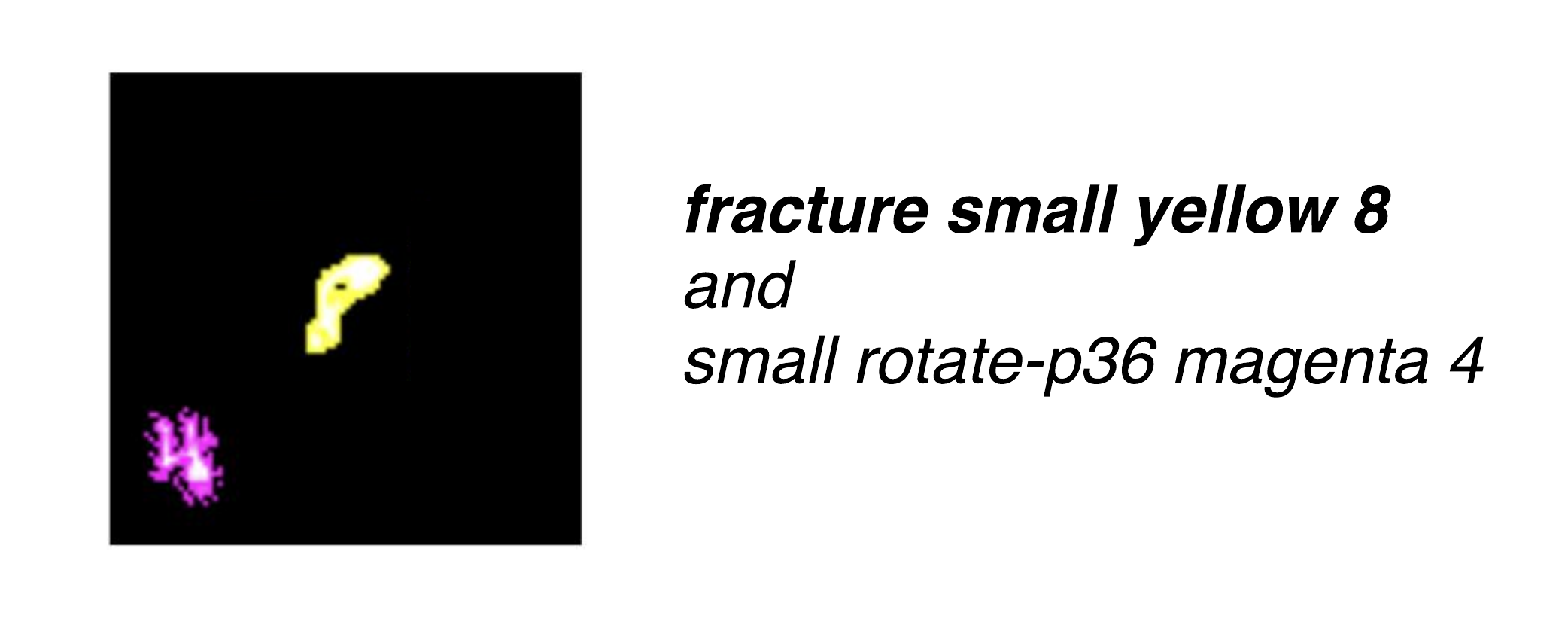}
    \caption[Illustration of saliency bias in MADMAN]{Illustration of saliency bias in MADMAN. Salient objects always occupy the center grid cell and are always included in the caption and receive priority placement (first).}
    \label{fig:saliency-bias-example}
\end{figure}

The MADMAN dataset explicitly models this phenomenon by implementing two strict criteria for salient objects: (1) fixed spatial positioning in the center cell of the image grid, and (2) guaranteed inclusion and primary position in the accompanying caption. See Fig. \ref{fig:saliency-bias-example} for visualization.

To systematically study the effects of saliency bias, the dataset generation process incorporates a saliency probability parameter $p_s$\label{notation:prob-saliency} that determines the proportion of examples exhibiting this bias pattern. This parameter enables controlled experiments examining how varying degrees of saliency bias influence model performance.

\begin{figure}[t!]
    \centering
    \includegraphics[width=\linewidth]{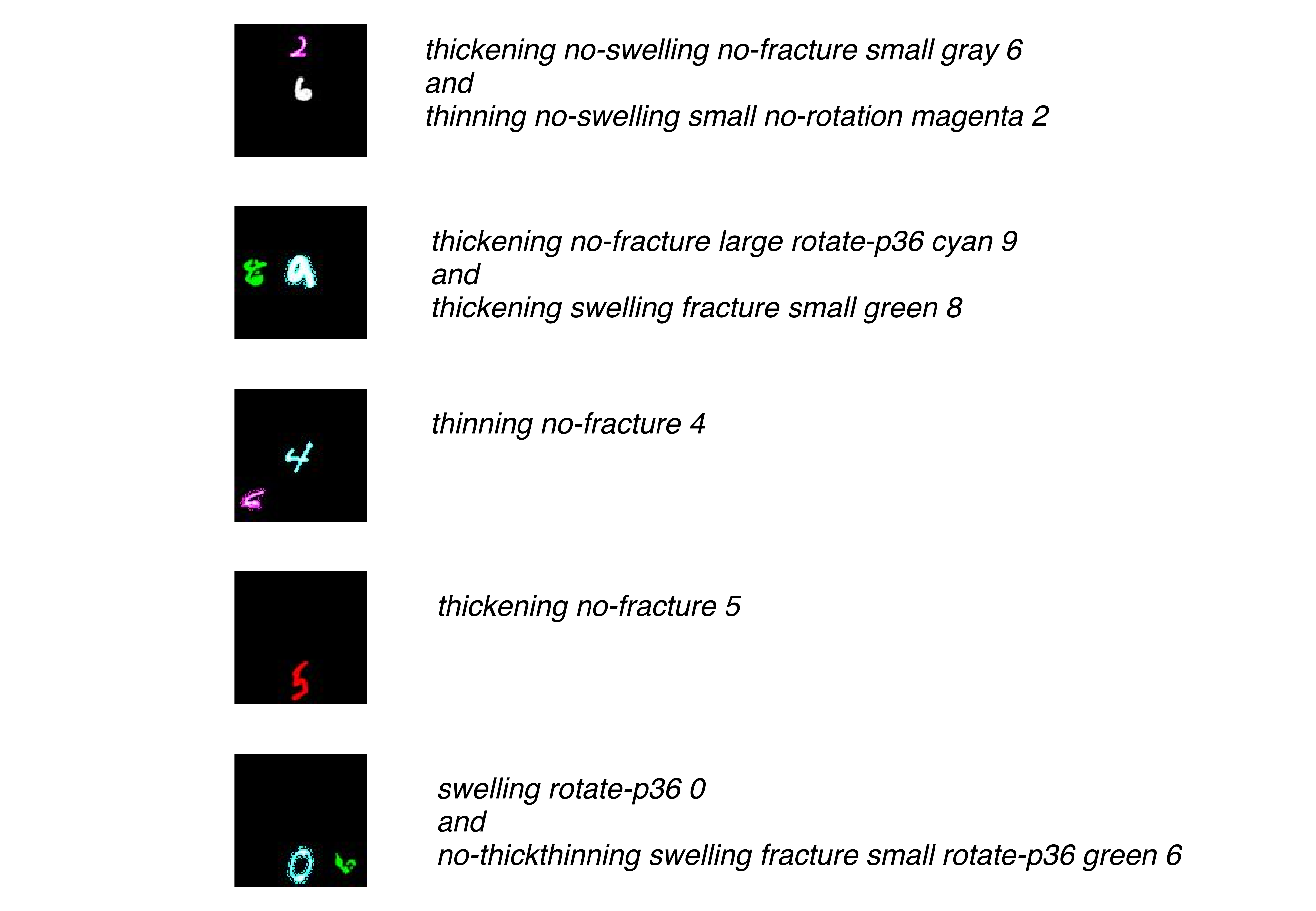}
    \caption[Examples in MADMAN]{Samples from the MADMAN dataset. The images are generated from transformed MNIST instances while the captions are created by chaining object (digit) name and the set of attributes (augmentations applied). There can be variable number of objects in the image, objects in the caption, and attributes described per object.}
    \label{fig:madman_example}
\end{figure}

\subsubsection{Model Details.}

The model is a scaled-down version of the standard CLIP design~\cite{clip} (with the ViT backbone). The reduction in model size and complexity reflects the simpler nature of our synthetic dataset, with its limited vocabulary, structured image composition, and controlled attribute variations.

\begin{table}[ht]
    \centering
    \begin{tabular}{lcc}
        \textbf{Parameter} & \textbf{TinyCLIP} & \textbf{clip-vit-base-patch32} \\
        embed\_dim         & 32                & 512                                   \\
        image\_resolution  & 96                & 224                                   \\
        vision\_layers     & 6                 & 12                                    \\
        vision\_width      & 48                & 768                                   \\
        vision\_n\_heads   & 4                 & 12                                    \\
        vision\_patch\_size & 7                 & 7                                     \\
        text\_encoder\_width & 32              & 512                                   \\
        context\_length    & 20                & 77                                    \\
        vocab\_size        & 34                & 49408                                 \\
        text\_n\_heads     & 4                 & 8                                     \\
        text\_layers       & 6                 & 12                                    \\
    \end{tabular}
    \caption[Comparison of architecture parameters between \texttt{TinyCLIP} and one of the original CLIP models]{Comparison of architecture parameters between our model \texttt{TinyCLIP}, used as the base model for the experiments, and \texttt{openai/clip-vit-base-patch32}. Due to the simple nature of the dataset we are working with (e.g. \texttt{vocab\_size} is three orders of magnitude smaller), we work with a smaller model. But as seen in \cref{validating_the_setup} the results also hold for a bigger model.}
\end{table}

\subsection{Finding Realistic values for the properties}
\label{realistic-value-procedure}

To estimate realistic values for the data properties, we manually annotated 100 samples (image-text pairs) from ConceptualCaptions12M~\cite{changpinyo2021cc12m} following the procedure below for each pair:

\begin{itemize}
    \item[-] \textbf{Count the number of objects in the image.} What counts as an object can be fuzzy but as a rule-of-thumb we took units that most humans may describe as being in the image if asked to be thorough. Note: because in the synthetic setup, we have one-object vs multi(two)-object, this counting doesn't need to very precise. From this, we estimated 0.95 as the proportion of multi-object images.
    \item[-] \textbf{Among the objects, count how many are included in the caption.} Once the objects are determined, we counted how many are mentioned in the caption. This is a conditional probability estimate. We then averaged the captioned/present ratio over all instance and determined the multi-object caption ratio to be around 0.6, which we took as the value for our realistic setup experiments.
    \item[-] \textbf{Count the number of attributes included in the caption for each object.} These only include objects identified in the first step. We found this distribution to be skewed with most objects having zero attributes defined. So we set the num-attributes distribution to the one with $\mathbb{E}[{n_a}]=0.57$ for our realistic setup.
    \item[-] \textbf{Determine whether one or more of the objects are salient.} Here, we consider objects to be salient if they standout visually for a human. We consider an image to have salient objects if there are any objects that are non-salient. From this we estimated saliency to be around 0.9 for a realistic setup.
\end{itemize}

\subsection{Performance in the Ideal Setups for each property}

Here we look at the performance when we take the base setup of our experiments in \cref{sec:results} and set each property separately to their ideal values. Because our base setup was already at the optimal values for three of the four properties (except \attrsInCap), we end up with identical and almost ideal results for most of the properties.

\label{individual_ideal_setup_performance}
\begin{table}[h]
    \centering
    \caption{Comparison of best setups found for individual attributes. Note that we performed our search by chance partially in the optimal setup, which results in identical results for the first three and last two rows.}
    \label{tab:comparison_best_setups}    
    \begin{tabular}{lccccccc}
        \toprule
        Setup & Color & Scaling & Fracture & Rotation & Swelling & Thickness  \\
        \midrule
        Best \multiObjRatioImg & 69.85 & 91.76 & 59.76 & 95.67 & 66.31 & 88.30  \\
        Best \multiObjRatioCap & 69.85 & 91.76 & 59.76 & 95.67 & 66.31 & 88.30 \\
        Best \saliencyBias & 69.85 & 91.76 & 59.76 & 95.67 & 66.31 & 88.30  \\
        Best \attrsInCap & 94.70 & 91.27 & 61.99 & 92.47 & 66.40 & 90.12  \\
        Combined Ideal data & 94.70 & 91.27 & 61.99 & 92.47 & 66.40 & 90.12  \\
        \bottomrule
    \end{tabular}
\end{table}

\subsection{Effect of Individual Realistic Factors in an Ideal Setting}
Here, we evaluate using the ideal setup but replacing each of the single factors with the realistic values as determined in \cref{realistic-value-procedure} i.e. \multiObjRatioCap=0.6, \multiObjRatioImg=0.95, \saliencyBias=0.9, and \attrsInCap with $\mathbb{E}[{n_a}]=0.57$.
\label{individual_realistic_sub}
\begin{table}[h]
    \centering
    \caption{Evaluation using ideal setup but replacing a single factor with a realistic value. Notably, performance drops significantly when Saliency Bias is set to a realistic level, even though all other factors remain ideal.}
    \label{tab:comparison_realistic_sub}    
    \resizebox{\linewidth}{!}{
    \begin{tabular}{lccccccc}
        \toprule
        Setup & Color & Scaling & Fracture & Rotation & Swelling & Thickness  \\
        \midrule
Full Ideal & 94.66 & 91.28 & 61.93 & 92.44 & 66.43 & 90.02 \\
        Ideal except \saliencyBias & 55.72 & 57.49 & 51.88 & 61.92 & 59.49 & 57.84 \\
        Ideal except \attrsInCap & 51.39 & 60.28 & 52.46 & 53.69 & 57.61 & 55.64 \\
        Ideal except \multiObjRatioImg & 83.51 & 93.02 & 60.89 & 93.96 & 66.62 & 91.87 \\
        Ideal except \multiObjRatioCap & 95.89 & 94.59 & 61.19 & 93.79 & 66.74 & 91.05 \\
Full Realistic & 50.47 & 54.47 & 52.34 & 50.86 & 53.26 & 50.05 \\
        \bottomrule
    \end{tabular}
    }
\end{table}

\newpage

\subsection{Recognition Performance in-distribution}
\label{recognition_in_distribution}

\begin{figure}[h]
    \centering
    \begin{subfigure}[b]{0.49\textwidth}
        \includegraphics[width=1\linewidth]{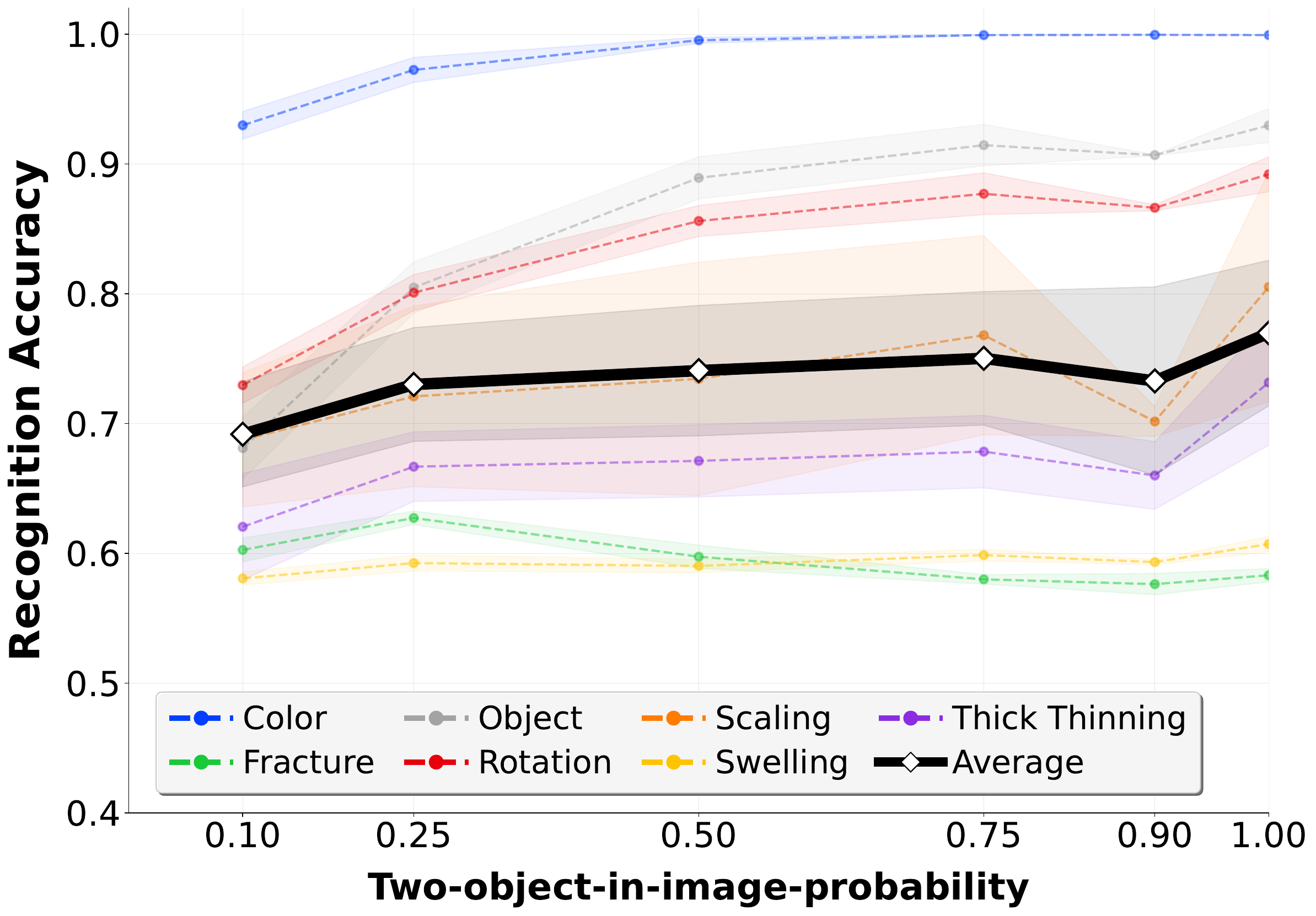}
        \caption{\textbf{\multiObjRatioImg}}
        \label{fig:test-in-recog-multi-object_image_ratio}
    \end{subfigure}
    \begin{subfigure}[b]{0.49\textwidth}
        \includegraphics[width=1\linewidth]{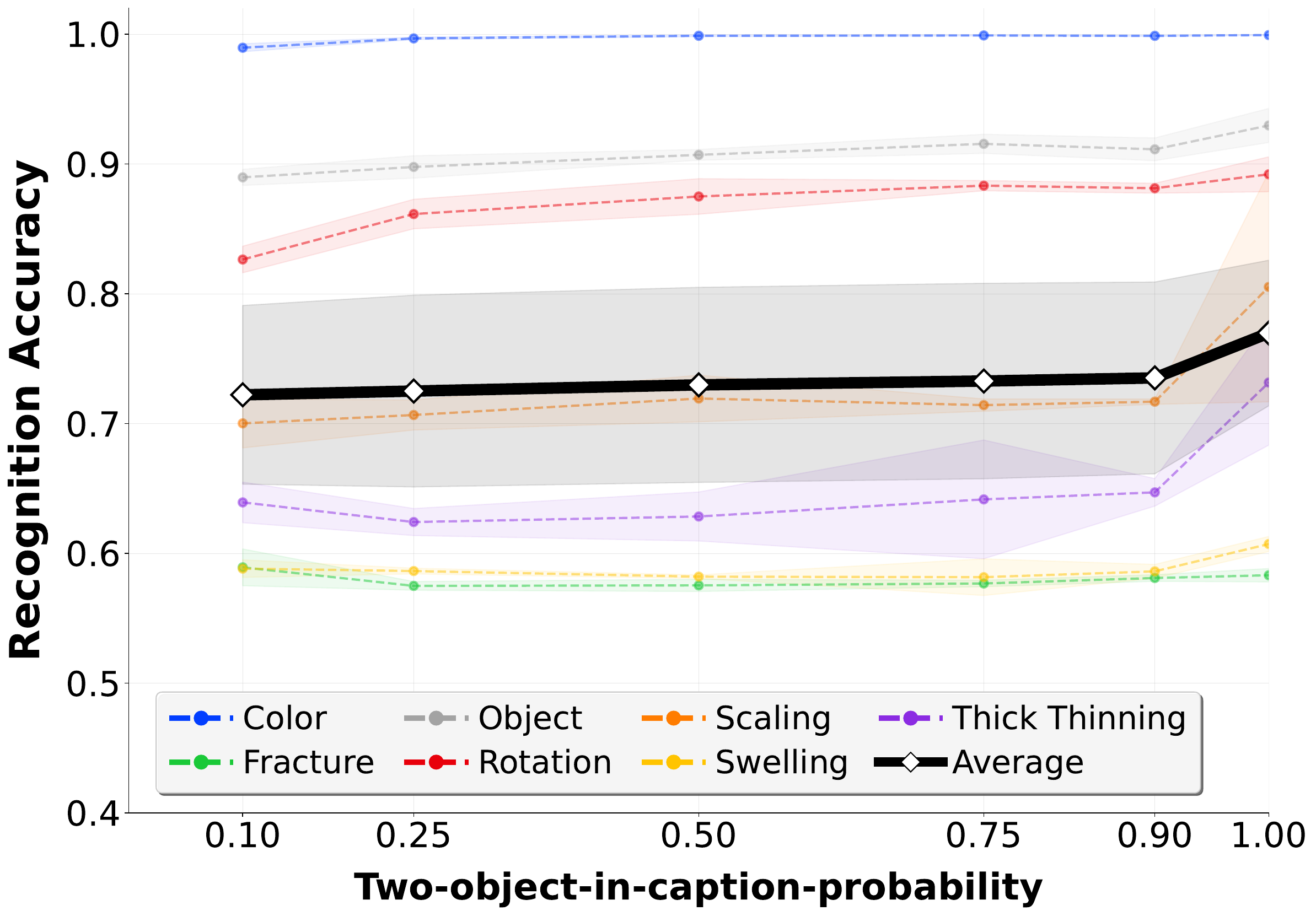}
        \caption{\textbf{\multiObjRatioCap}}
        \label{fig:test-in-recog-multi-object_caption_ratio}
    \end{subfigure}
    
    \begin{subfigure}[b]{0.49\textwidth}
        \includegraphics[width=1\linewidth]{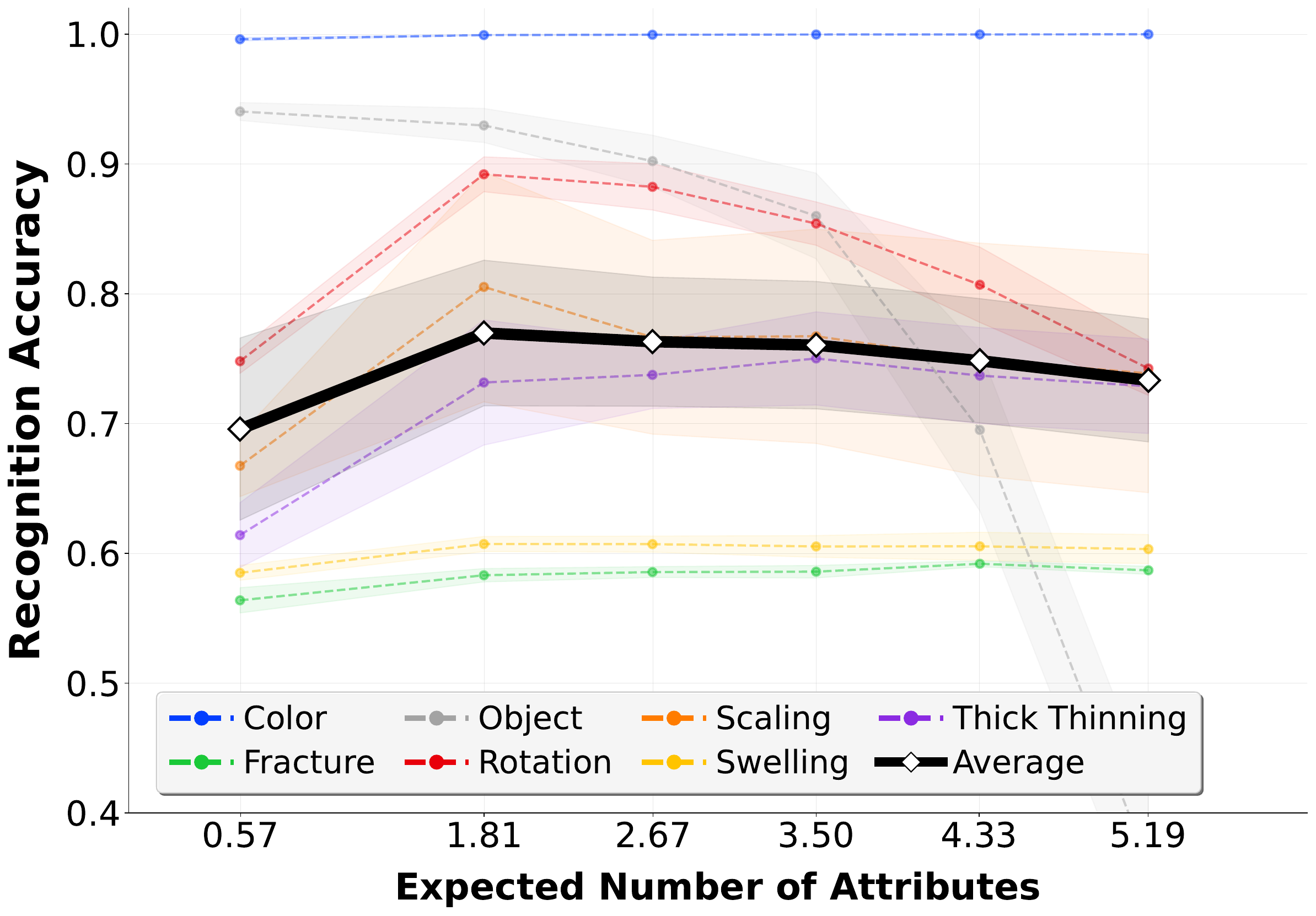}
        \caption{\textbf{\attrsInCap}}
        \label{fig:test-in-recog-nr_attr_in_cap}
    \end{subfigure}
    \begin{subfigure}[b]{0.49\textwidth}
        \includegraphics[width=1\linewidth]{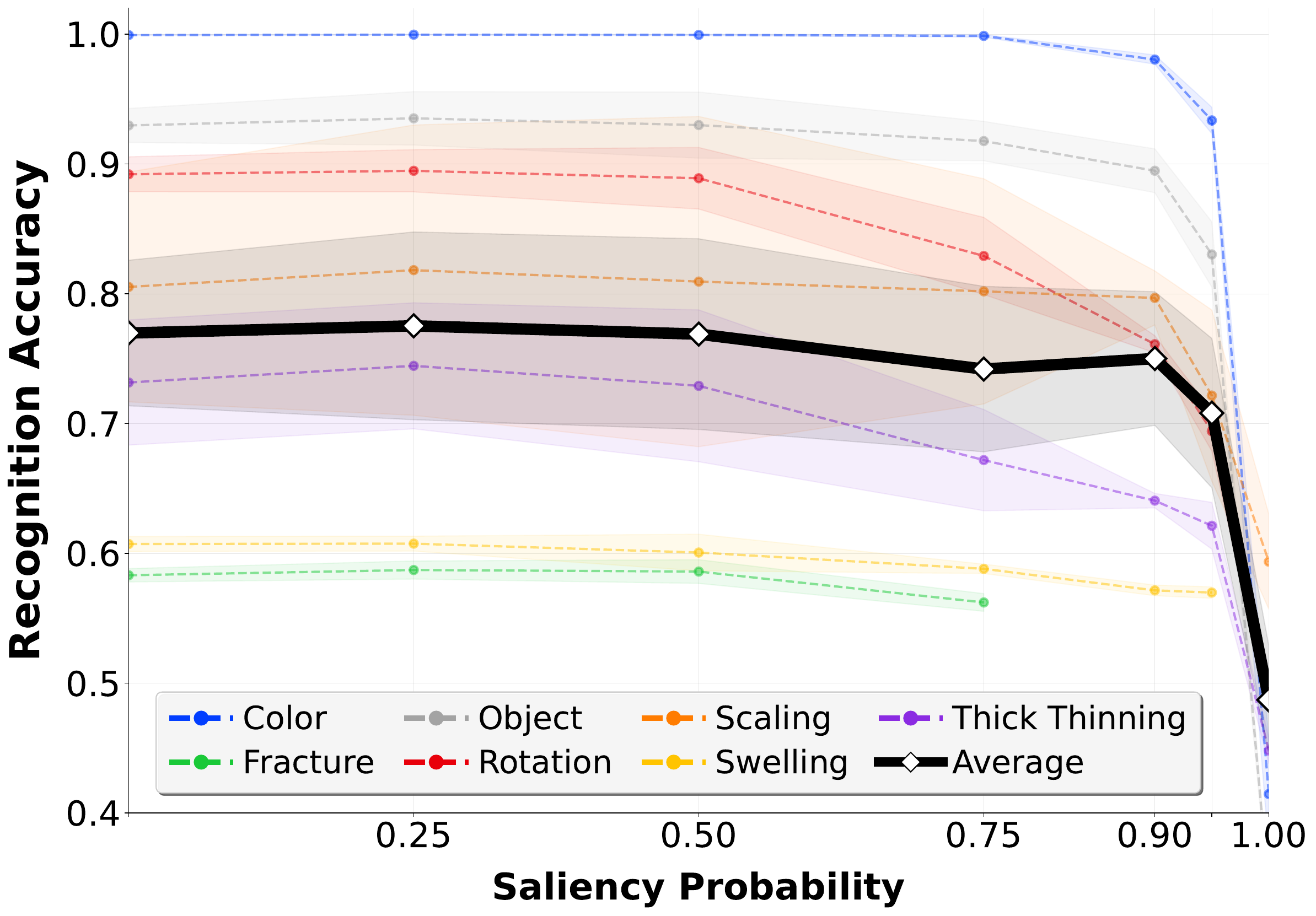}
        \caption{\textbf{\saliencyBias}}
        \label{fig:test-in-recog-saliency}
    \end{subfigure}
    
    \caption{Impact of the data properties on the attribute recognition for in-distribution combinations. On average, it remains stable throughout, although the levels are different for the different attributes. Notably object-recognition drops for high number of attributes in the caption, likely because the model can rely more on color to distinguish samples in the batch as its present with a higher probability.}
    \label{fig:in-distribution-recognition-results}
\end{figure}

\newpage
\subsection{Recognition Performance out-of-distribution}
\label{recog-ood-all}

\begin{figure}[ht]
    \centering
    \begin{subfigure}[b]{0.49\textwidth}
        \includegraphics[width=1\linewidth]{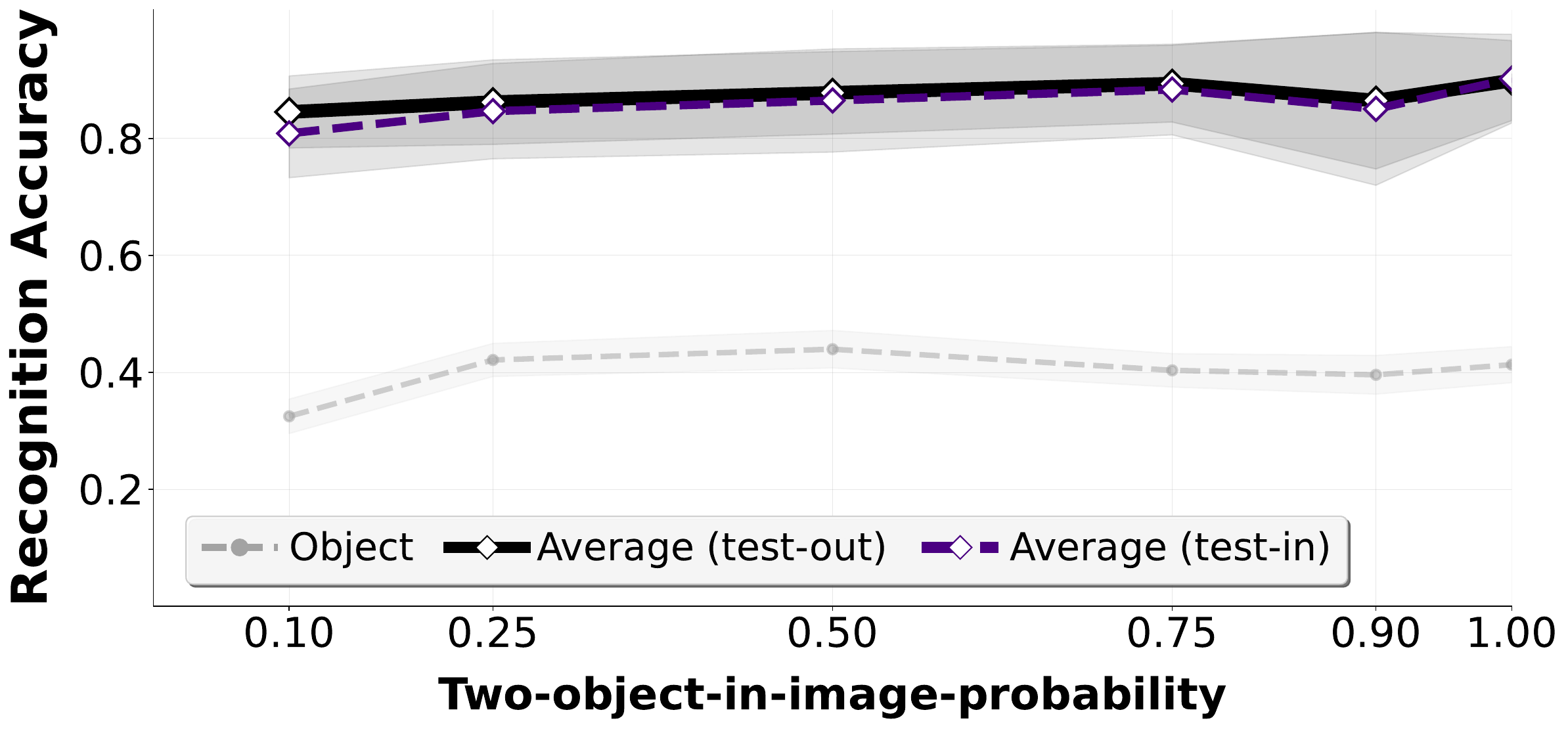}
        \caption{\textbf{\multiObjRatioImg}}
        \label{fig:test-out-recog-multi-object_image_ratio}
    \end{subfigure}
    \begin{subfigure}[b]{0.49\textwidth}
        \includegraphics[width=1\linewidth]{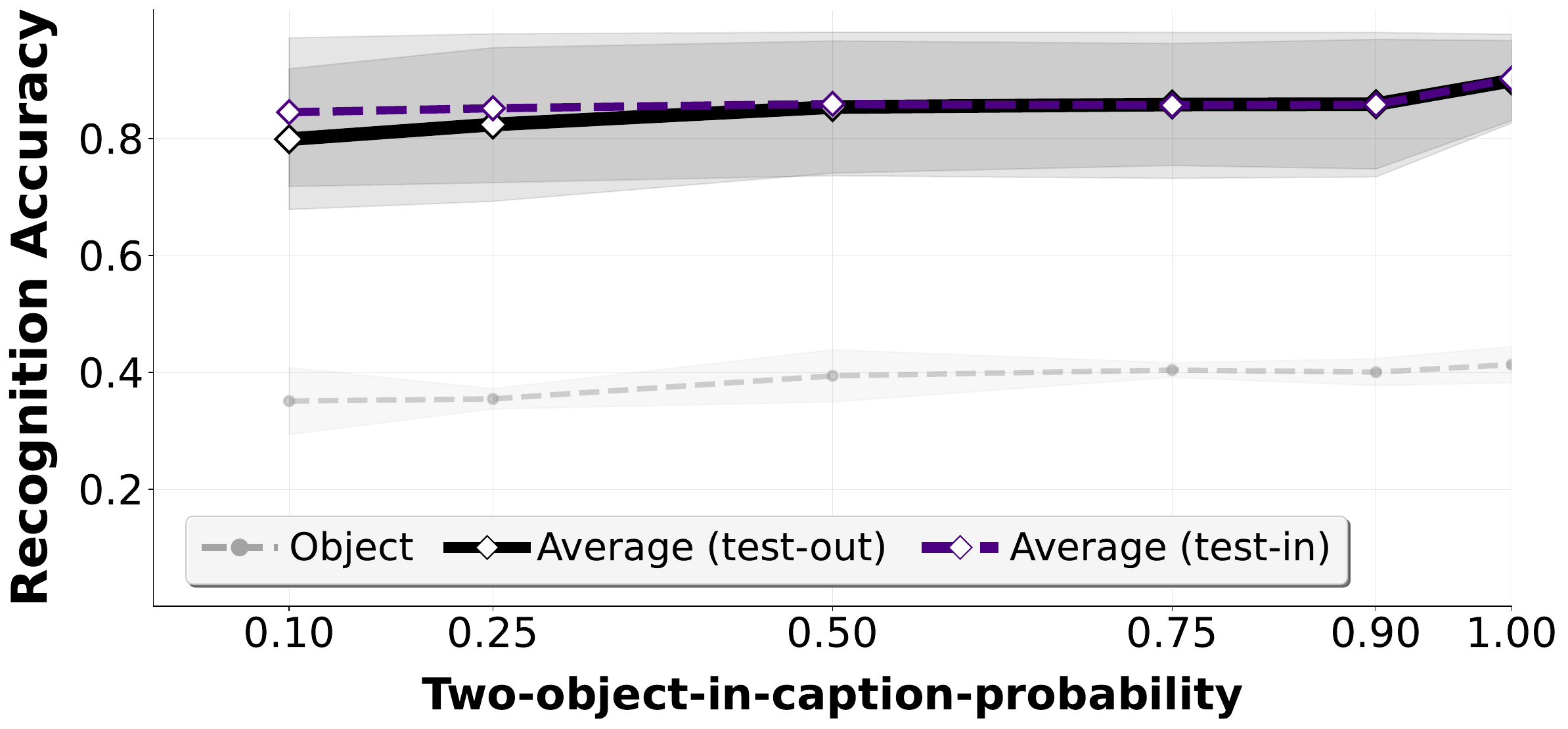}
        \caption{\textbf{\multiObjRatioCap}}
        \label{fig:test-out-recog-multi-object_caption_ratio}
    \end{subfigure}
    
    \vspace{0.5em} 

    \begin{subfigure}[b]{0.49\textwidth}
        \includegraphics[width=1\linewidth]{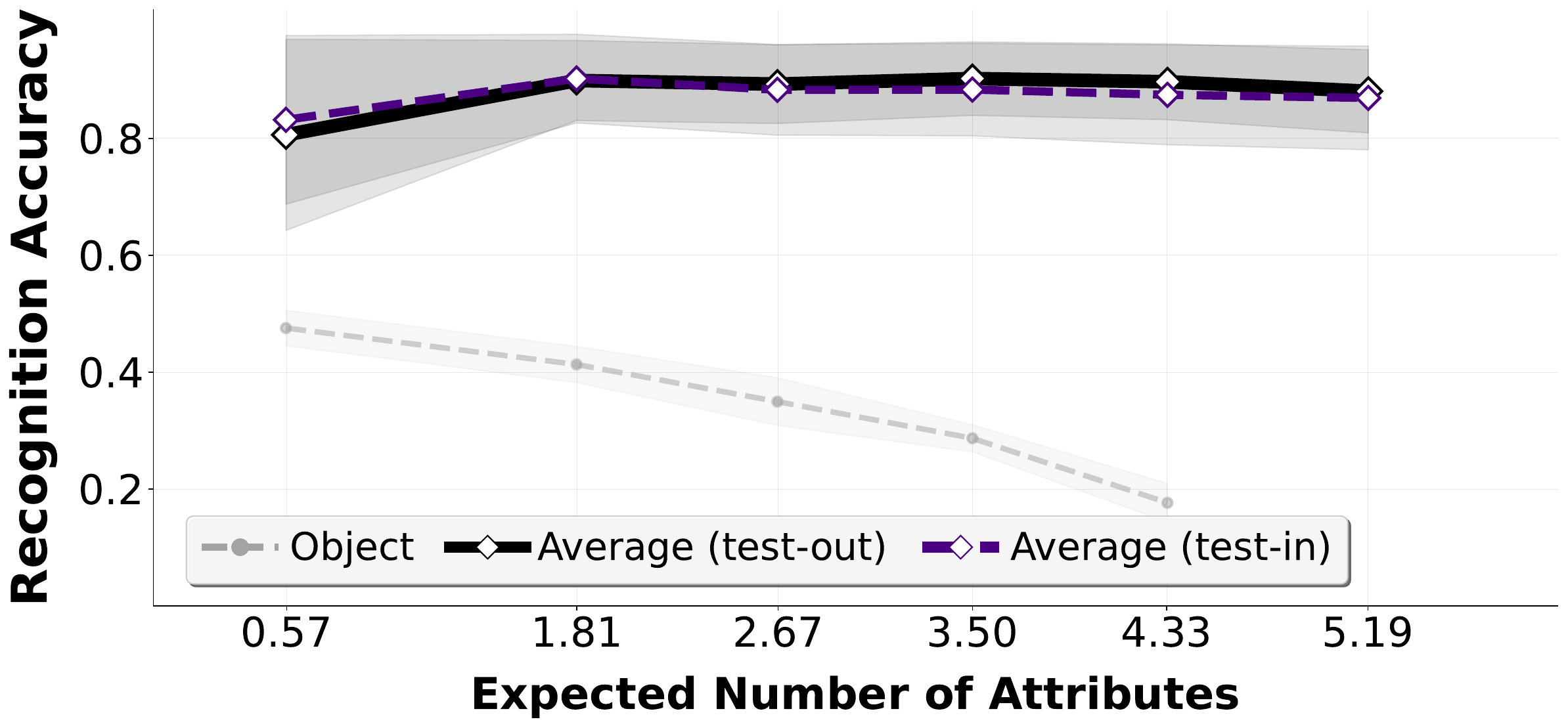}
        \caption{\textbf{\attrsInCap}}
        \label{fig:test-out-recog-nr_attr_in_cap}
    \end{subfigure}
    \begin{subfigure}[b]{0.49\textwidth}
        \includegraphics[width=1\linewidth]{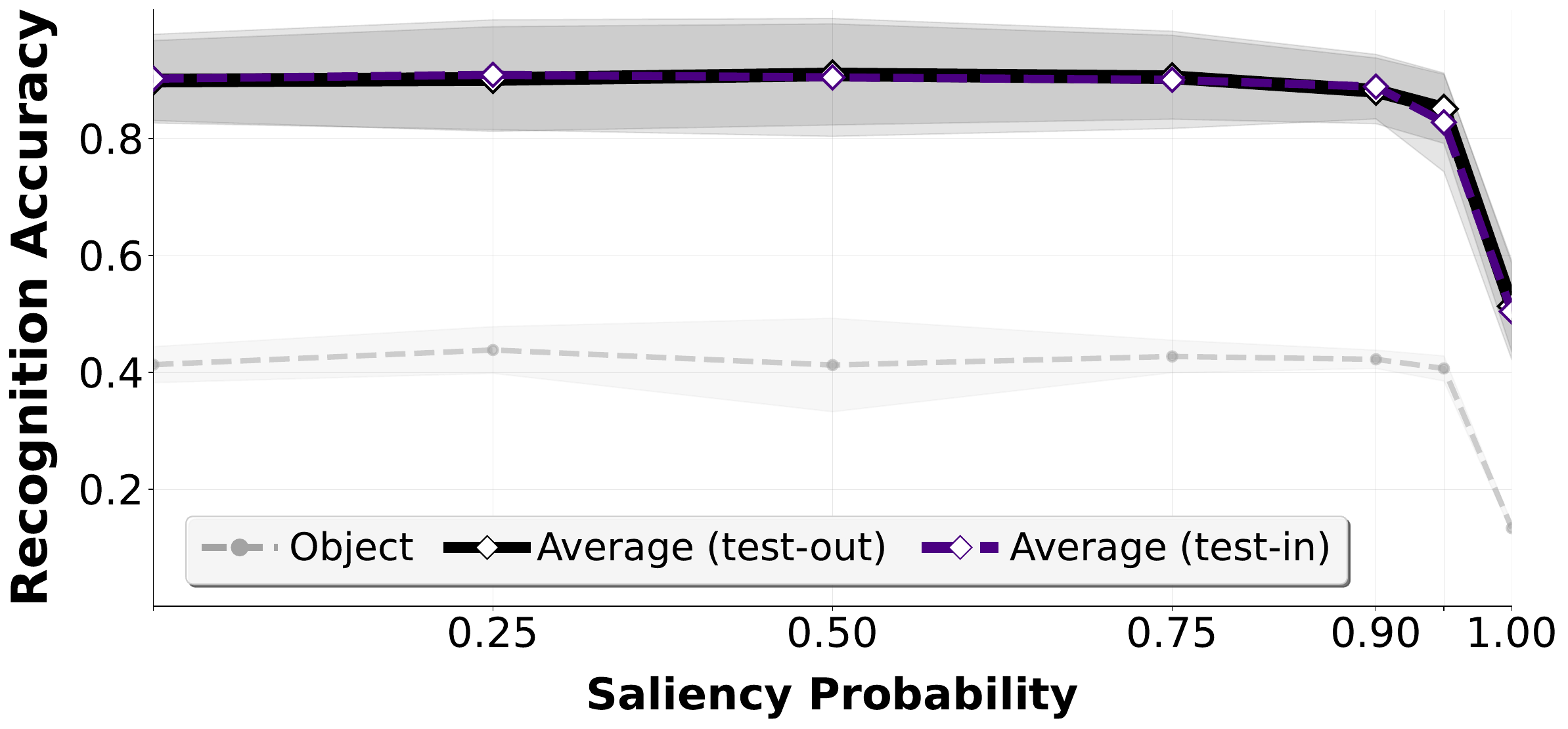}
        \caption{\textbf{\saliencyBias}}
        \label{fig:test-out-recog-saliency}
    \end{subfigure}
    
    \caption{Impact of the data properties on the attribute recognition for out-of-distribution combinations. It follows similar trends as in-distribution (denoted by the dotted average). However, the object-recognition is generally lower.}
    \label{fig:out-of-distribution-recognition-results}
\end{figure}

\newpage

\subsection{Influence of Data Properties on Object-Attribute Binding (No Recognition Filtering)}

\begin{figure}[h]
    \centering
    \begin{subfigure}[b]{0.49\textwidth}
    
        \includegraphics[width=1\linewidth]{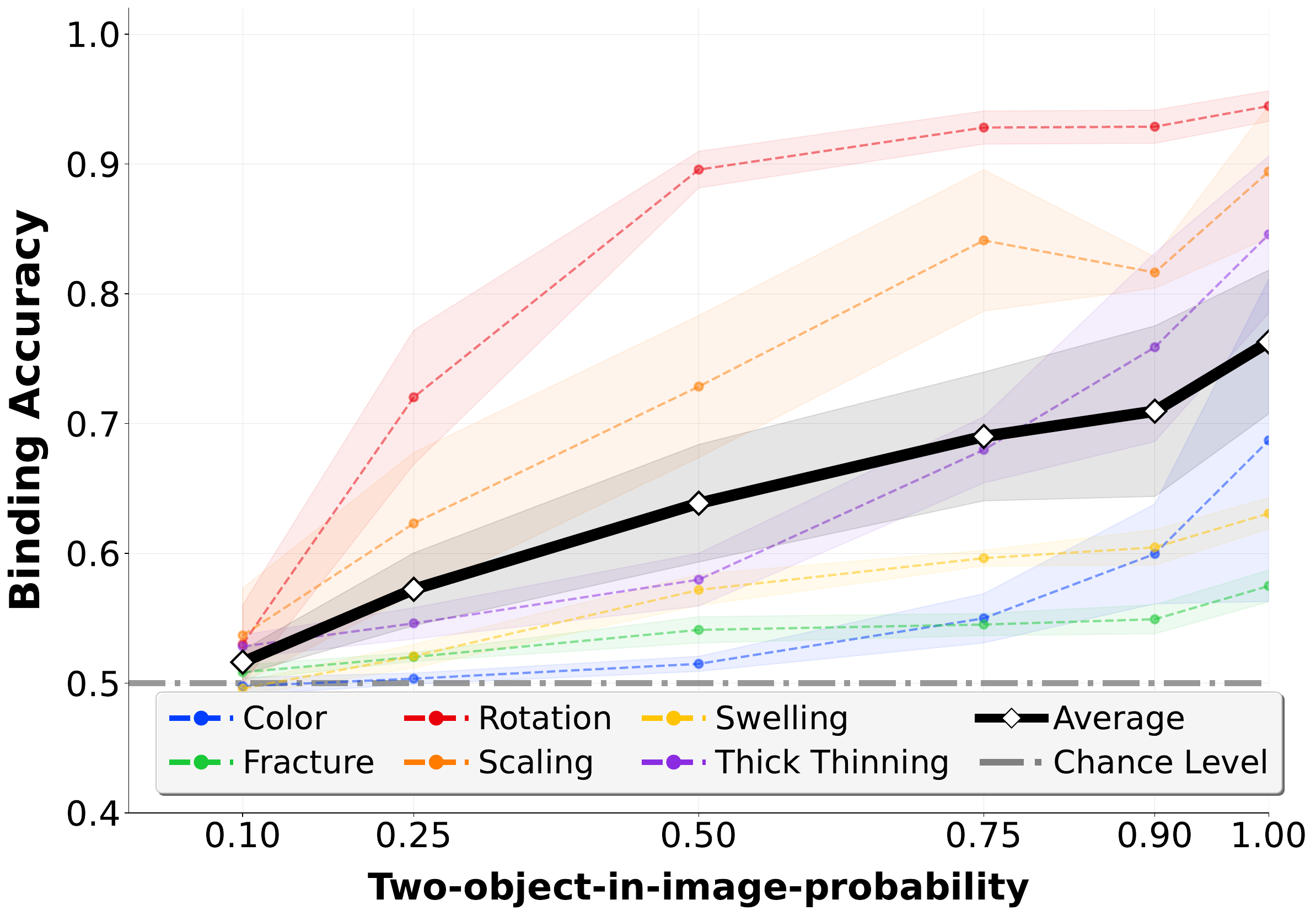}
        \caption{\textbf{\multiObjRatioImg}}
        \label{fig:multi-object_image_ratio_uncond}
    \end{subfigure}    
    \begin{subfigure}[b]{0.49\textwidth}    
        \includegraphics[width=1\linewidth]{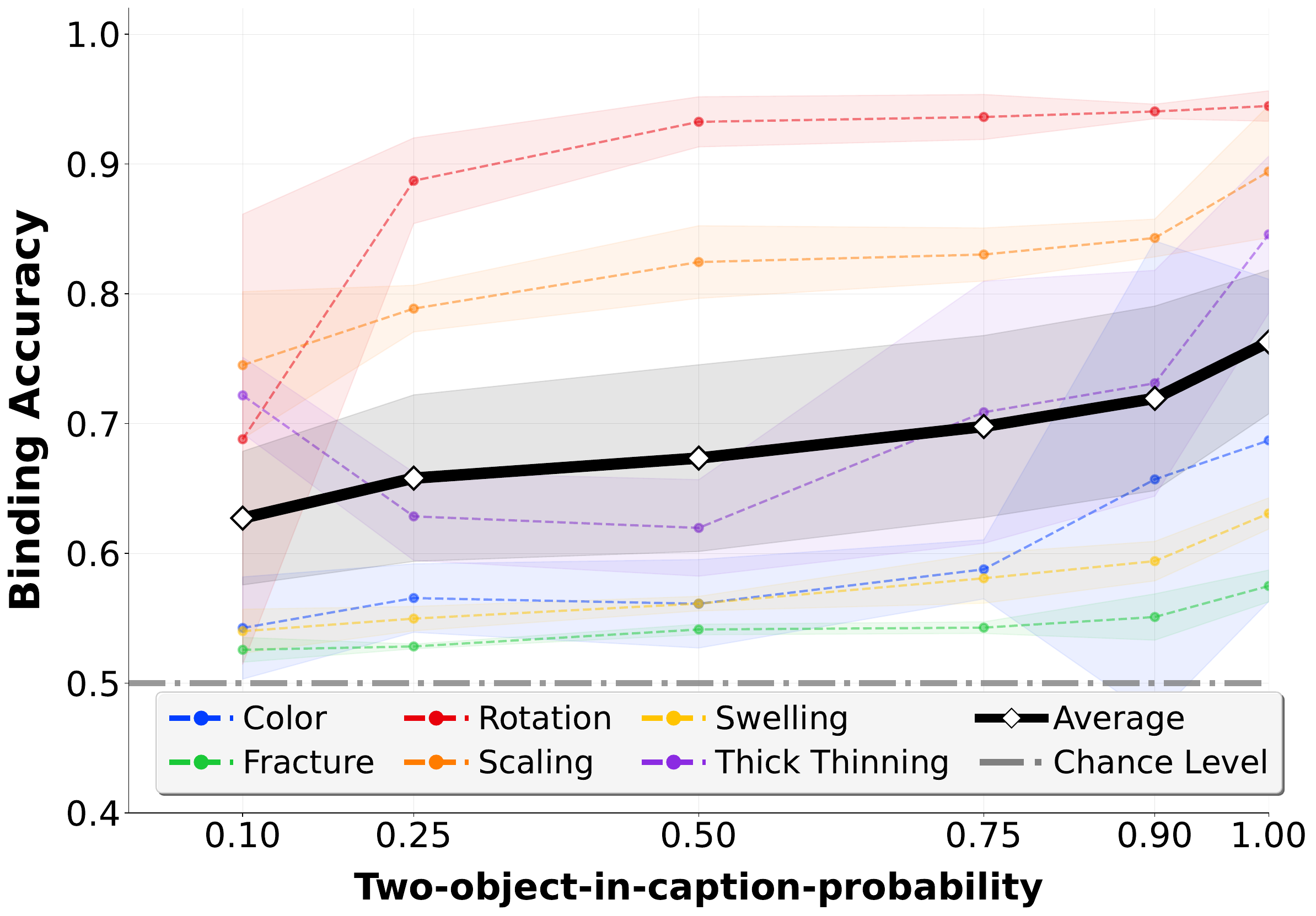}
        \caption{\textbf{\multiObjRatioCap}}
        \label{fig:multi-object_caption_ratio_uncond}
    \end{subfigure}    
    \caption{\textbf{Influence of (a) \multiObjRatioImg and (b) \multiObjRatioCap on binding accuracy on MADMAN, evaluated without recognition filtering.}
    The trend here is the same i.e. having more images and captions containing multiple objects improves binding.}
    \label{fig:multi-object_ratios_uncond}
\end{figure}

\begin{figure}[t]
    \centering
    \begin{subfigure}[b]{0.49\textwidth}
        \includegraphics[width=1\linewidth]{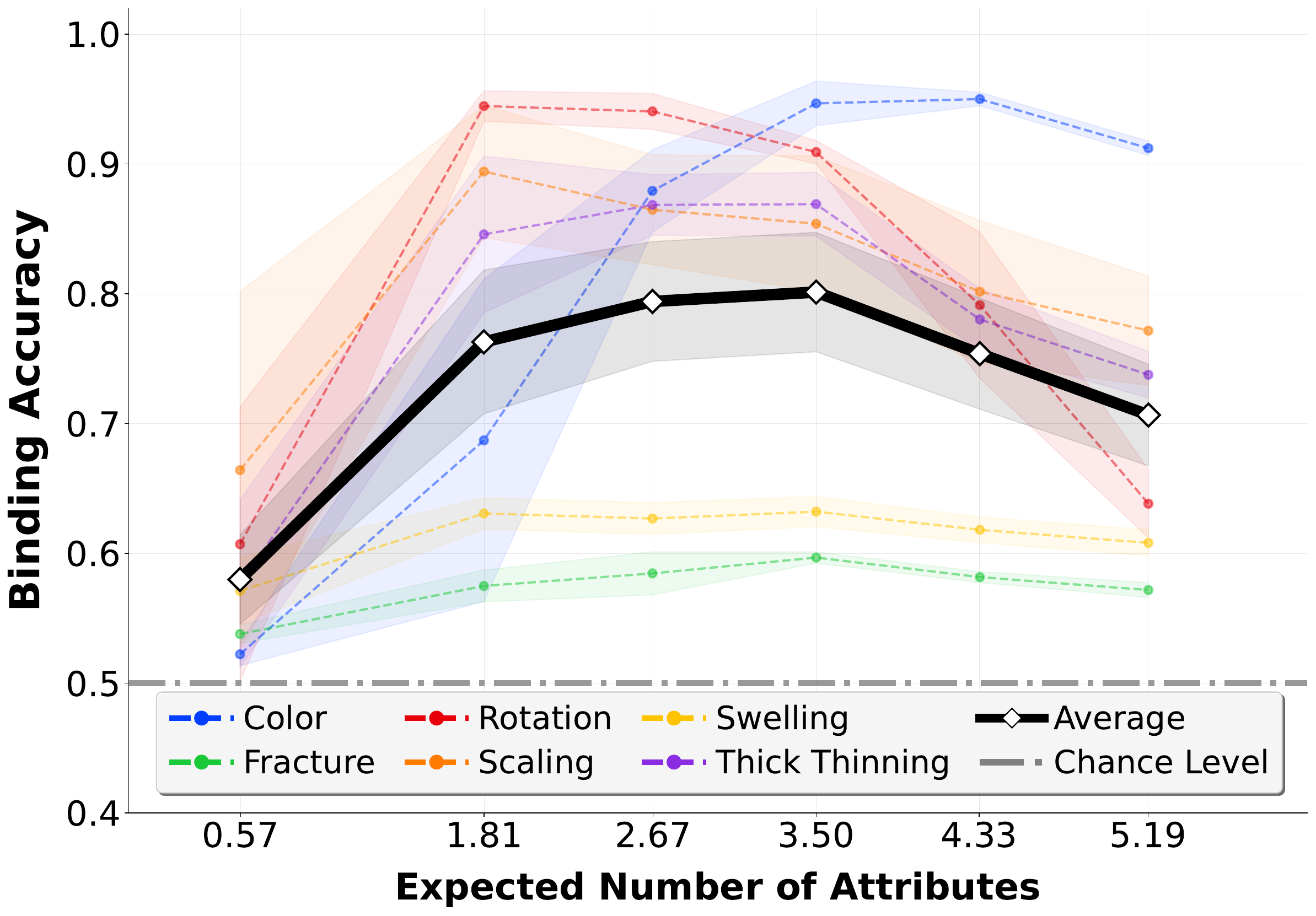}
        \caption{\textbf{\attrsInCap}}
        \label{fig:nr_attr_in_cap_uncond}
    \end{subfigure}    
    \begin{subfigure}[b]{0.49\textwidth}    
        \includegraphics[width=1\linewidth]{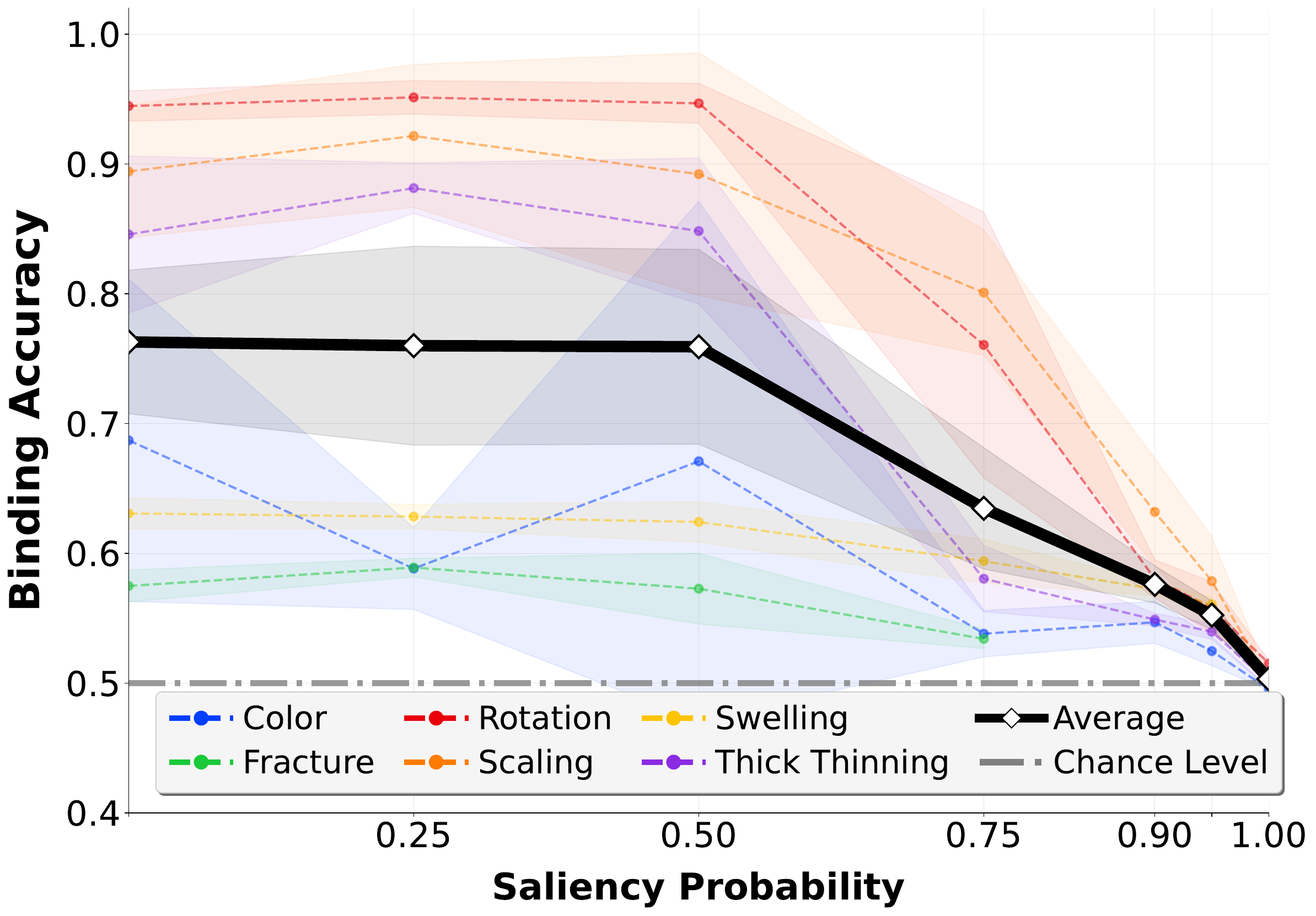}
        \caption{\textbf{\saliencyBias}}
        \label{fig:saliency_uncond}
    \end{subfigure}    
    \caption{
    \textbf{Influence of (a) \attrsInCap and (b) \saliencyBias on binding accuracy on MADMAN evaluated without recognition filtering.}
    It follows the same trend as in \cref{fig:num-attrs-saliency} i.e. too few and too many \attrsInCap hamper the binding. And high \saliencyBias is also detrimental to the binding}
    \label{fig:num-attrs-saliency_uncond}
\end{figure}

In \cref{sec:what-limits} we looked at the impact of the data properties under consideration on the object-attribute binding. For the results there, as detailed in \cref{eval-and-training-setup}, we filter out samples where the attribute is not recognized for both the objects.
This filtering makes sense as it factors out binding failures caused simply by failed recognitions. However, it introduces a slight variation in the test set between runs, since each evaluation uses a slightly different subset of instances. To ensure that this filtering does not significantly affect the overall trend, we also present results without recognition-based filtering. As seen in \cref{fig:multi-object_ratios_uncond} and \cref{fig:num-attrs-saliency_uncond}, the trend of how each of the properties influences the binding stays the same.

\end{document}